\documentclass[sigconf, nonacm]{acmart}

\usepackage{multirow}
\usepackage{multicol}
\usepackage{tabu}
\usepackage{graphicx}
\usepackage{makecell}
\usepackage{amsthm}
\usepackage{mathtools}
\usepackage{microtype}
\usepackage[export]{adjustbox}
\usepackage[T1]{fontenc}
\usepackage[scaled=0.825]{beramono}
\usepackage{hhline}
\usepackage{xspace}
\usepackage[ruled,vlined]{algorithm2e}
\usepackage{lipsum}
\usepackage{comment}
\usepackage[super]{nth}
\usepackage{array}
\usepackage{balance}
\usepackage{bm}
\newcolumntype{P}[1]{>{\centering\arraybackslash}p{#1}}
\newcolumntype{L}[1]{>{\arraybackslash}p{#1}}
\usepackage{hhline}

\usepackage{amsmath,amsfonts}
\usepackage{algorithmic}
\usepackage{textcomp}
\usepackage{xcolor}
\usepackage{listings, color}
\usepackage{tikz}
\usepackage{comment}

\usepackage{hyperref}

\newcommand{\tablesize}{\fontsize{7.5pt}{7.5pt}\selectfont}
\newcommand{\para}[1]{\noindent \textbf{#1.}}

\newcommand{\rev}[1]{\textcolor{black}{#1}}

\definecolor{mauvelous}{rgb}{0.94, 0.6, 0.67}
\definecolor{mauvetaupe}{rgb}{0.57, 0.37, 0.43}
\lstdefinestyle{myStyle}{
  lineskip=0mm,
  belowcaptionskip=1\baselineskip,
  frame=tb,
  language=python,
  aboveskip=0mm,
  belowskip=0mm,
  showstringspaces=false,
  columns=flexible,
  basicstyle={\fontsize{7pt}{7pt}\fontfamily{fvm}\selectfont},
  numbers=left,
  xleftmargin=3em,
  numberstyle={\color{gray}\texttt},
  keywordstyle=\color{black}\textbf,
  commentstyle=\color{mauvelous},
  frame=none,
  breaklines=true,
  breakatwhitespace=true,
  tabsize=2,
  morekeywords={where},
  keywords={[2]{}},
}

\newcommand{\name}{Ginex\xspace}
\newcommand{\first}{\texttt{superbatch} \texttt{sample}\xspace}
\newcommand{\second}{\texttt{changeset} \texttt{precomputation}\xspace}
\newcommand{\third}{\texttt{feature} \texttt{cache} \texttt{initialization}\xspace}
\newcommand{\fourth}{\texttt{main} \texttt{loop}\xspace}

\newcommand\vldbdoi{10.14778/3551793.3551819}
\newcommand\vldbpages{2626 - 2639}
\newcommand\vldbvolume{15}
\newcommand\vldbissue{11}
\newcommand\vldbyear{2022}
\newcommand\vldbauthors{\authors}
\newcommand\vldbtitle{\shorttitle} 
\newcommand\vldbavailabilityurl{https://github.com/SNU-ARC/Ginex}
\newcommand\vldbpagestyle{empty}

\usepackage{multirow}
\usepackage{adjustbox}
\usepackage{listings}
\usepackage{tikz}
\usepackage{amsmath}

\newcommand*\circled[1]{\tikz[baseline=(char.base)]{
            \node[shape=circle,fill,inner sep=0.7pt] (char) {\textcolor{white}{#1}};}}

\microtypecontext{spacing=nonfrench}

\begin{document}

\title{\name : SSD-enabled Billion-scale Graph Neural Network Training on a Single Machine via Provably Optimal In-memory Caching}

\author{Yeonhong Park}
\affiliation{%
  \institution{Seoul National University}
  \city{Seoul}
  \country{Korea}
}
\email{ilil96@snu.ac.kr}

\author{Sunhong Min}
\orcid{0000-0002-2247-232X}
\affiliation{%
  \institution{Seoul National University}
  \city{Seoul}
  \country{Korea}
}
\email{sunhongmin@snu.ac.kr}


\author{Jae W. Lee}
\orcid{0000-0001-5109-3700}
\affiliation{%
  \institution{Seoul National University}
  \city{Seoul}
  \country{Korea}
}
\email{jaewlee@snu.ac.kr}

\begin{abstract}
Graph Neural Networks (GNNs) are receiving a spotlight as a powerful tool that can effectively serve various inference tasks on graph structured data.  As the size of real-world graphs continues to scale, the GNN training system faces a scalability challenge. Distributed training is a popular approach to address this challenge by scaling out CPU nodes. However, not much attention has been paid to \emph{disk-based} GNN training, which can scale up the single-node system in a more cost-effective manner by leveraging high-performance storage devices like NVMe SSDs. We observe that the data movement between the main memory and the disk is the primary bottleneck in the SSD-based training system, and that the conventional GNN training pipeline is sub-optimal without taking this overhead into account. Thus, we propose \name, the first SSD-based GNN training system that can process billion-scale graph datasets on a single machine. Inspired by the inspector-executor execution model in compiler optimization, \name restructures the GNN training pipeline by separating \emph{sample} and \emph{gather} stages. This separation enables \name to realize a provably optimal replacement algorithm, known as \emph{Belady's algorithm}, for caching feature vectors in memory, which account for the dominant portion of I/O accesses. According to our evaluation with four billion-scale graph datasets and two GNN models, \name achieves 2.11$\times$ higher training throughput on average (2.67$\times$ at maximum) than the SSD-extended PyTorch Geometric.

\end{abstract}
\maketitle

\pagestyle{\vldbpagestyle}
\begingroup\small\noindent\raggedright\textbf{PVLDB Reference Format:}\\
\vldbauthors. \vldbtitle. PVLDB, \vldbvolume(\vldbissue): \vldbpages, \vldbyear.\\
\href{https://doi.org/\vldbdoi}{doi:\vldbdoi}
\endgroup
\begingroup
\renewcommand\thefootnote{}\footnote{\noindent
This work is licensed under the Creative Commons BY-NC-ND 4.0 International License. Visit \url{https://creativecommons.org/licenses/by-nc-nd/4.0/} to view a copy of this license. For any use beyond those covered by this license, obtain permission by emailing \href{mailto:info@vldb.org}{info@vldb.org}. Copyright is held by the owner/author(s). Publication rights licensed to the VLDB Endowment. \\
\raggedright Proceedings of the VLDB Endowment, Vol. \vldbvolume, No. \vldbissue\ %
ISSN 2150-8097. \\
\href{https://doi.org/\vldbdoi}{doi:\vldbdoi} \\
}\addtocounter{footnote}{-1}\endgroup

\ifdefempty{\vldbavailabilityurl}{}{
\vspace{.3cm}
\begingroup\small\noindent\raggedright\textbf{PVLDB Artifact Availability:}\\
The source code, data, and/or other artifacts have been made available at \url{\vldbavailabilityurl}.
\endgroup
}

\section{Introduction}
\label{sec:intro}

Recently, the success of Deep Neural Network (DNN) has extended its scope of application to graphs beyond images and texts. As a new class of DNN, Graph Neural Network (GNN) is now proving itself to be a powerful tool~\cite{survey1, survey2, survey3} that can replace the traditional graph analytic methods in various inference tasks on graph-structured data, such as node classification~\cite{node-classification}, recommendation~\cite{pinsage, pinner-sage},  and link prediction~\cite{link-prediction}. Owing to its expressive power, GNN effectively captures the rich relational information embedded among input nodes, leading to decent generalization performance.

Meanwhile, the GNN training process features unique challenges in its data preparation stage. In GNN, unlike in traditional DNNs where data samples are independent of each other (e.g., images), the nodes in a graph are closely connected with each other. To perform a single iteration of mini-batch GNN training, we need feature vectors of not only the nodes in the mini-batch but also their neighbor nodes~\cite{gcn-1,node-classification}.
This first requires finding neighbors of the nodes in the mini-batch by traversing the graph, and then collecting their sparsely located feature vectors in a contiguous buffer for the next steps. These two operations, called \texttt{sample} and \texttt{gather}, respectively, involve a large number of data accesses. 

For this reason, existing popular GNN frameworks~\cite{dgl2019wang,pyg} opt to keep the whole graph dataset in the main memory throughout the training process. Disk-based GNN training has not received much attention so far due to concerns on its performance~\cite{gnnssd}.
However, there is yet an imperative to explore disk-based GNN training further because of the prevalence of gigantic graph datasets with billions of nodes and edges. The size of real-world graph datasets reaches hundreds of GB or even a few TB (and growing), which may exceed the main memory capacity~\cite{aligraph, agl}. Several works have previously addressed the scalability issue of in-memory GNN training by having more CPU nodes~\cite{p32021ghandi,aligraph2019zhu,agl,cm-gcn,dist-dgl,dist-dglv2}. However, a significant increase in system cost can limit the effectiveness of this approach since they scale all hardware components by the same factor, even if some of them may be underutilized. Instead, disk-based GNN training is a promising alternative as modern NVMe SSDs can offer enough capacity to hold the entire input graph as well as much higher read performance than the previous generations.

Thus, we propose \name (\underline{G}raph \underline{in}spector-\underline{ex}ecutor), the first GNN training system based on high-performance commodity NVMe SSDs. While being cost-effective for capacity scaling, SSD-based GNN training system is obviously challenging for an order of magnitude lower bandwidth than DRAM and lack of byte-addressability. Therefore, it is important to reduce the amount of I/O requests as much as possible. To this end, \name introduces a technique to effectively utilize the main memory space as in-memory cache. Especially, for \texttt{gather} which is the most I/O intensive job in GNN training, \name realizes the optimal caching mechanism. Inspired by the inspector-executor execution model~\cite{inex-1, inex-2} in compiler optimization, \name reorganizes the GNN training process into two phases. In the first phase which corresponds to \emph{inspector}, \name performs \texttt{sample} for a large enough number of batches. It allows preparation of the optimal caching mechanism for the following \texttt{gather} operations by analyzing the sampling results. In the second phase, which corresponds to \emph{executor}, \name completes the remaining jobs for the batches including \texttt{gather} with the cache managed by the guidance derived in the first phase. 

We prototype \name on PyTorch Geometric (PyG)~\cite{pyg}, a popular library for GNN. We evaluate \name on a server with an 8-core Intel Xeon CPU with 64GB memory and a NVIDIA V100 GPU with 16GB memory, using four billion-scale graph datasets whose total size ranges from 326GB to 569GB. Our evaluation shows that \name reduces the training time by 2.11$\times$ on average (2.67$\times$ at maximum) compared to PyG extended to support disk-based processing of graph dataset with a reasonable additional storage overhead. In addition, our case study on Google Cloud demonstrates that single-node \name saves the training cost by 2.76$\times$ and 5.71$\times$ compared to 8-node DistDGL~\cite{dist-dgl}, a popular distributed GNN training system, for \textit{Friendster} and \textit{Twitter} datasets, respectively. 

The followings are the summary of our contributions:
\begin{itemize}
    \item We profile SSD-based GNN training and make two main observations regarding the sub-optimality of the OS page cache and the conventional training pipeline.
    \item Based on our observations, we introduce a novel GNN training pipeline that enables realization of the optimal caching mechanism for feature gathering, which cannot be realized in the conventional training pipeline.
    \item We present an efficient algorithm to simulate the optimal cache replacement policy and accelerate it with GPU.
    \item We prototype \name on PyG, a popular GNN library, and demonstrate its effectiveness using four billion-scale datasets that do not fit in memory.
\end{itemize}

\section{Background}
\label{sec:back}

\subsection{Graph Neural Networks}
\label{sec:back:gnn}
\begin{figure}[t]
  \centering
  \includegraphics[width=\linewidth]{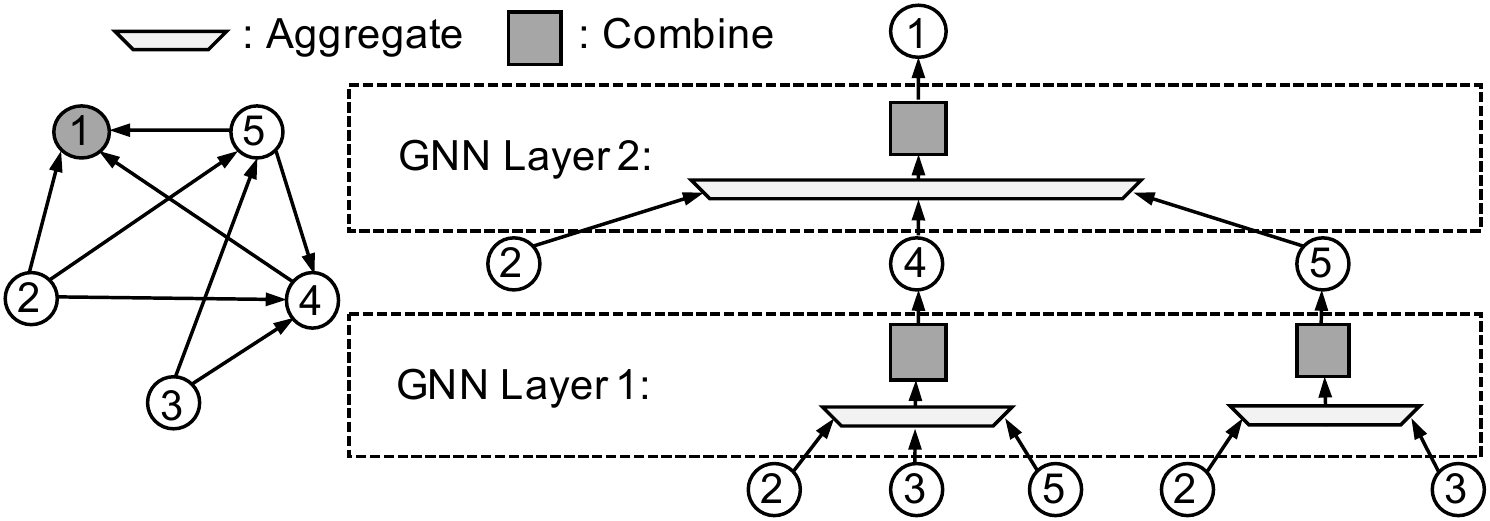}
  \caption{2-layer GNN training on Node 1}
  \label{fig:gnn-example}
\end{figure}
\para{GNN Training} A GNN operates on graph-structured data, where each node has its own feature vector. GNNs aim to produce a quality embedding for each node in the graph capturing its neighborhood information on top of its own feature.
These embeddings may be used for various downstream tasks such as node classification and link prediction. To obtain the embedding of a node, GNN takes the feature vectors of not only the target node for embedding computation, which is called \emph{seed node}, but also its \emph{k}-hop in-neighbors as input.
Each layer in GNN is responsible for synthesizing feature information of the nodes at each hop, which means that \emph{k}-layer GNN is able to reflect up to \emph{k}-hop in-neighbors~\cite{gcn-1,node-classification}. 

Each layer of GNN consists of two main steps: \texttt{Aggregate} and \texttt{Combine}. The embedding of node $v$ after the $i$th layer, denoted as $h^{i}_{v}$, is computed as the following:
\begin{equation}
h^{i}_{v} = \texttt{Combine}(\texttt{Aggregate}(\left\{ h^{i-1}_{u} \mid u\in N(v)\right\}))
\end{equation}
$N(v)$ denotes the neighbor set of node $v$. In \texttt{Aggregate} step, the features of the incoming nodes are aggregated into a single vector.
While popular options for aggregation functions are simple operations like mean, max and sum, more sophisticated aggregation functions are also drawing attention~\cite{gat}.
The aggregated feature then goes through \texttt{Combine} step which is essentially a fully connected (FC) layer with a non-linear function. Figure~\ref{fig:gnn-example} illustrates this process with an example of a 2-layer GNN training on Node $1$.

\para{Neighborhood Sampling}
\begin{figure}[t]
  \centering
  \includegraphics[width=\linewidth]{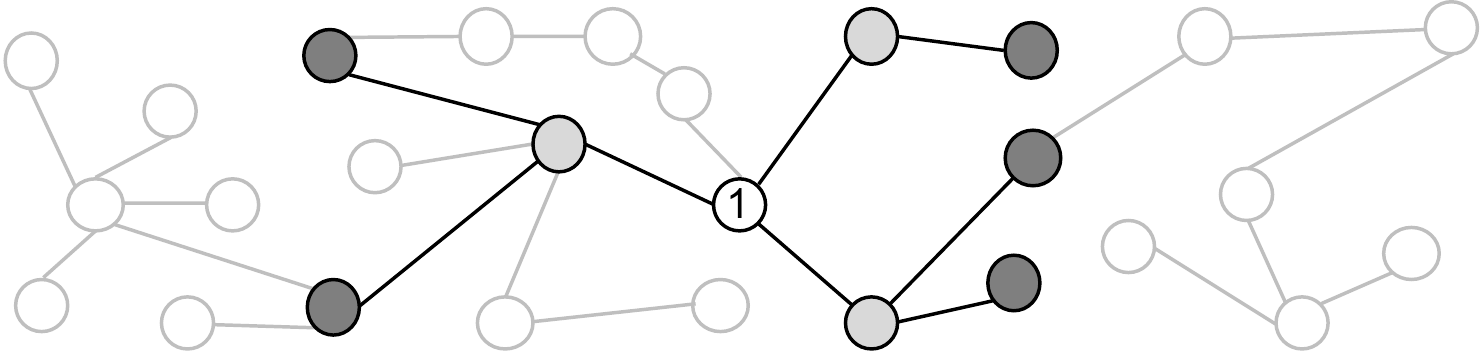}
  \caption{Sampling for a 2-layer GNN (sampling size = (3,2), batch size = 1)}
  \label{fig:sampling}
\end{figure}
The inter-node dependence of training data poses a unique challenge to GNN training. Even if we use small batch size, the training cost for each batch can still be quite high because collecting $k$-hop in-neighbors leads to exponential growth of memory footprint. Neighborhood sampling is a popular technique for this \emph{neighborhood explosion} problem. Instead of sampling $k$-hop in-neighbors of seed nodes, sampling algorithms select only a subset of them. One representative work is GraphSAGE~\cite{graphsage}, which randomly samples only a predefined number of in-neighbors at each aggregation step. Figure~\ref{fig:sampling} shows an example with a 2-hop computational graph for Node $1$ being sampled.
The sampling size in this example is (3,2) which means that it selects (at most) three among the neighbor nodes connected to the target node (Node $1$) and (at most) two are selected for each of the previously selected nodes. Its variants differ in several aspects of sampling function design like the granularity of sampling operation or the choice of probability distribution for sampling~\cite{vrgcn,mvs,pinsage,fastgcn}. In practice, it is not usual to go beyond three layers, and popular choices of sampling size for GraphSAGE are (25, 10), (10, 10, 10), and (15, 10, 5)~\cite{pytorch-direct}. 

\subsection{GNN Training System}
\label{sec:back:pipeline}
\begin{figure}[t]
  \centering
  \includegraphics[width=\linewidth]{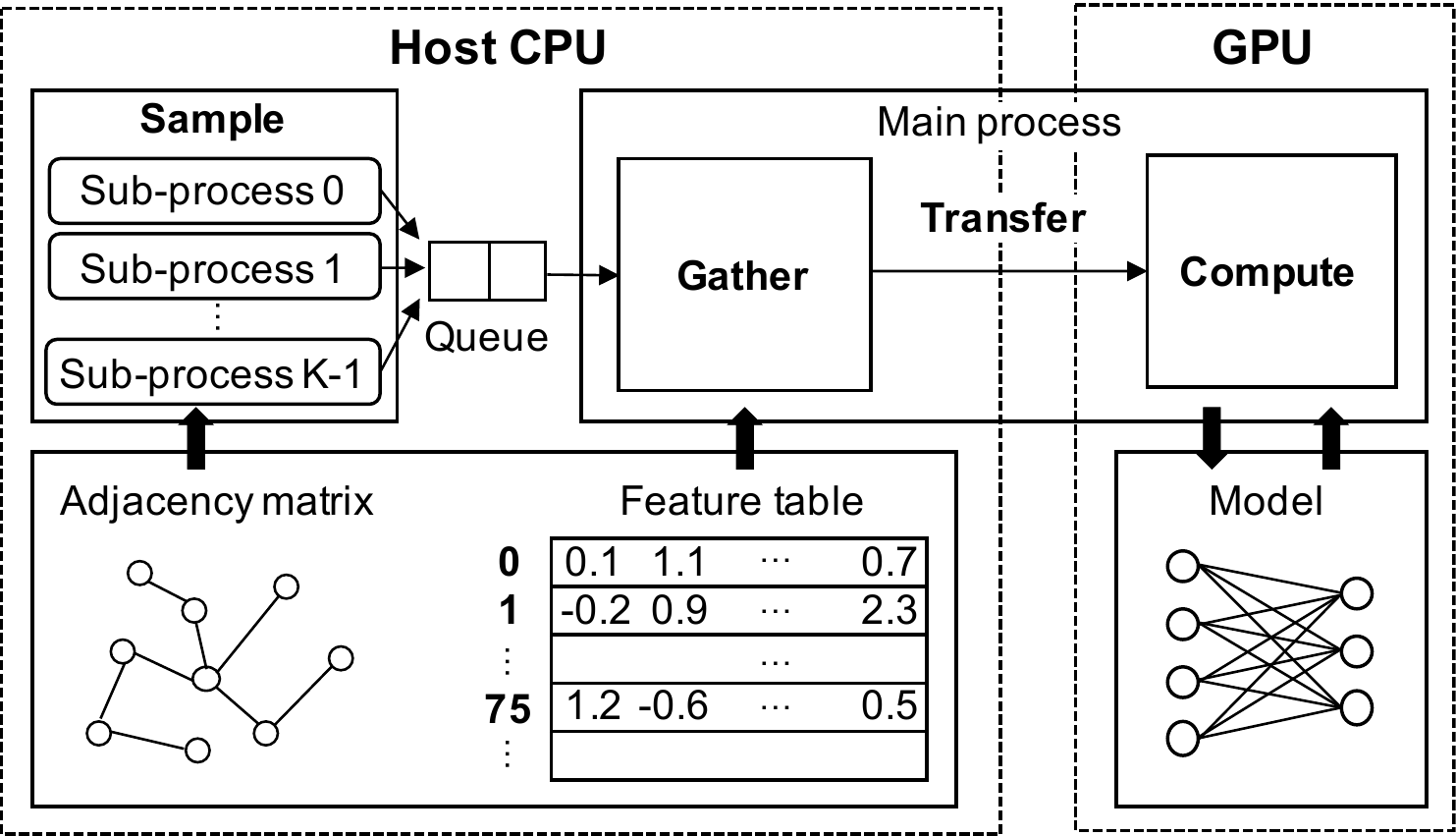}
  \caption{Overview of conventional GNN training system}
  \label{fig:pipeline}
\end{figure}
The state-of-the-art DNN frameworks~\cite{dgl2019wang, pyg} employ a mixed CPU-GPU training system, where CPU stores the graph data and is in charge of data preparation, whereas GPU executes the core GNN operations, i.e., \texttt{aggregate} and \texttt{combine}. GPU memory capacity is often fairly limited to store the graph data, while the massive parallelism of GPU is key to accelerating GNN computations. Figure~\ref{fig:pipeline} visualizes a typical process of mixed CPU-GPU training of GNN which consists of four steps: (1) \texttt{sample}, (2) \texttt{gather}, (3) \texttt{transfer}, and (4) \texttt{compute}. At every iteration, seed nodes for a single batch as well as their neighbors are extracted by traversing the graph structure (i.e., adjacency matrix) (\texttt{sample}). The adjacency matrix is usually stored in the compressed sparse column (CSC) format as it allows fast access to in-neighbors of each node. Then, the sparsely located feature vectors of the sampled nodes are collected into a contiguous buffer (\texttt{gather}), which is transferred to GPU over PCIe interface (\texttt{transfer}). Lastly, GPU performs forward/backward propagation to compute gradients and updates the parameters (\texttt{compute}). Since sampling operation is often memory-intensive, it is common to spawn multiple sub-processes to increase the sampling throughput. Each sub-process performs a sampling job for a batch, and puts the result into a shared queue. The main process fetches the sampling result from the shared queue and executes the remaining jobs, i.e., \texttt{gather}, \texttt{transfer}, and \texttt{compute}.

\subsection{Disk-based GNN Training}
\label{sec:ssd-based-training}
\para{Need for Disk-based GNN Training}
Although the size of the largest publicly available GNN dataset, \textit{ogbn-papers100M}~\cite{ogb}, is around 100GB, companies reportedly operate on much larger internal datasets that have hundreds of millions or even billions of nodes and tens of billions of edges~\cite{aligraph, agl}. For these datasets, the size of both node feature table and adjacency matrix may reach several hundreds of GBs or even a few TBs, often exceeding the memory capacity of a single node. Distributed training which partitions the graph dataset into multiple nodes in a cluster is a popular solution. However, scaling-out is not necessarily the most cost-effective solution to scale memory capacity~\cite{buri}. In this case, instead of adhering to in-memory processing, leveraging high-performance storage devices like NVMe SSDs as memory extension can be a promising direction owing to its large capacity and hence cost efficiency~\cite{gnnssd}.



\label{sec:back:ssd-based-training}

\para{Bottleneck Analysis}
However, it is still difficult to achieve high throughput on an SSD-based GNN training system. SSD operates as a block device where data is transferred in a 4KB chunk while \texttt{sample} and \texttt{gather} usually consist of fine-grained random accesses whose size ranges from tens to hundreds of bytes. Such a coarse access granularity combined with its relatively low bandwidth results in huge I/O penalty. In fact, we have profiled the execution time of SSD-based GNN training. We have trained GraphSAGE with a sampling size of (10,10,10) on a 8-core Intel Xeon CPU with 64GB memory and an NVIDIA V100 GPU equipped with 16GB HBM2 memory. We pipelined the execution of not only \texttt{sample} but also \texttt{gather} with the CUDA operations (\texttt{transfer} and \texttt{compute}) by creating a separate thread for \texttt{gather}.
Two large-scale graph datasets whose size of both adjacency matrix and feature table far exceed the host memory capacity are synthetically generated by extending the real world datasets (\textit{ogbn-papers100M}, \textit{ogbn-products})~\cite{gnnssd}.
We denote \textit{ogbn-papers100M} and \textit{ogbn-products} by \textit{papers} and \textit{products}, respectively, in following figures.
The graph dataset (i.e., adjacency matrix and feature table) is memory-mapped by \texttt{mmap} syscall. The details on the setup and the datasets are elaborated in Section~\ref{sec:eval:setup}. 

\begin{figure}[t]
  \centering
  \includegraphics[ width=\linewidth]{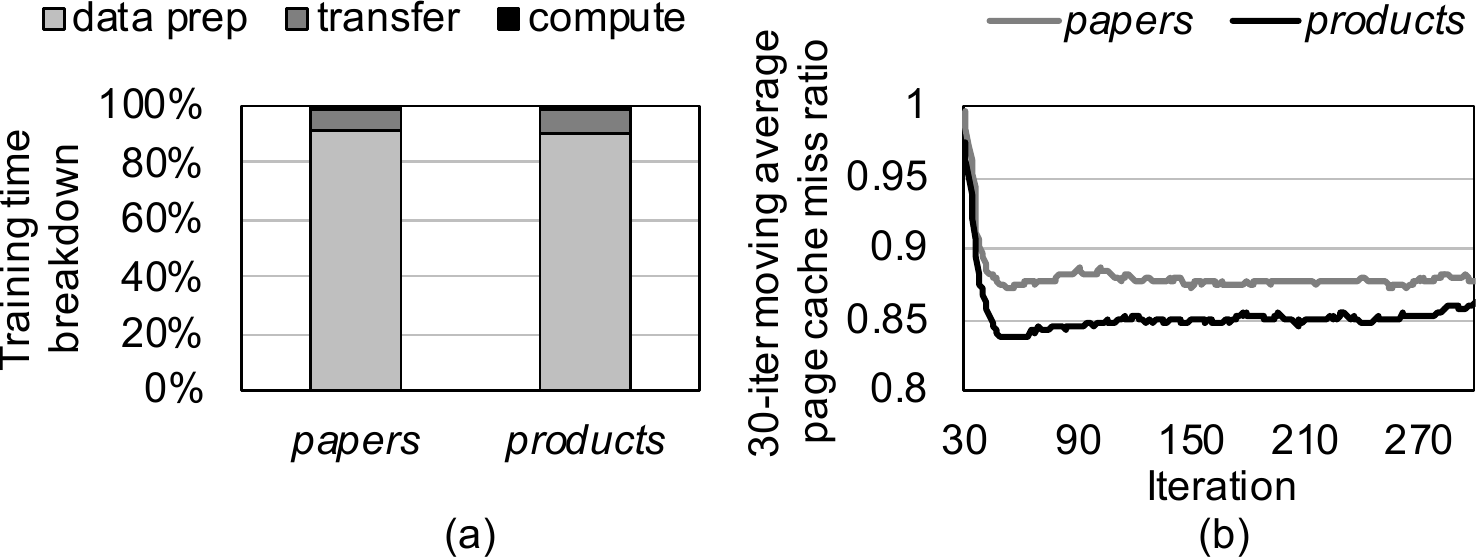}
  \caption{(a) Training time breakdown and (b) OS page cache miss ratio of SSD-based GNN training system with memory-mapped graph dataset}
  \label{fig:motiv-profiling}
\end{figure}

Figure~\ref{fig:motiv-profiling}(a) shows the training time breakdown, and Figure~\ref{fig:motiv-profiling}(b) the page cache \textit{miss} ratio of the GNN training program over time by measuring a 30-iteration moving average.
For both datasets, more than 90\% of training time is stalled by data preparation (i.e., \texttt{sample} and \texttt{gather}). In addition, the page cache miss ratio is fairly high. The miss ratio rapidly drops in the beginning as the page cache gets filled, but it soon stabilizes at 85-90\%. This implies that I/O is the bottleneck, and thus it is important to reduce the number of I/Os in \texttt{sample} and \texttt{gather}. Among the two, especially \texttt{gather} needs to be optimized. This is because the number of I/Os is usually several-fold higher in \texttt{gather} than in \texttt{sample}. Assuming a 3-layer model, all the sampled 3-hop neighbors of the seed nodes should be gathered for each iteration. Meanwhile, examining neighborhood information of the 2-hop neighbors of the seed nodes is enough to sample the 3-hop neighbors. In fact, in our experiments, the number of I/O requests made in \texttt{gather} is reported to be $8.13\times$ and $9.75\times$ higher than in \texttt{sample} for \textit{ogbn-papers100M} and \textit{ogbn-products}, respectively. 

\para{Challenge 1. Sub-optimal In-memory Cache}
As OS page cache, which simply keeps the recently accessed pages, is shown to be ineffective for GNN training, application-specific in-memory cache can be a viable alternative. There have been proposals to cache part of a graph dataset considering the access pattern of GNN training. Although they usually assume different training scenarios and target to optimize different types of data movement like communications among cluster machines or between CPU and GPU, not disk I/O, their spirits can be extended for the SSD-based training system.
\begin{figure}[t]
  \centering
  \includegraphics[ width=\linewidth]{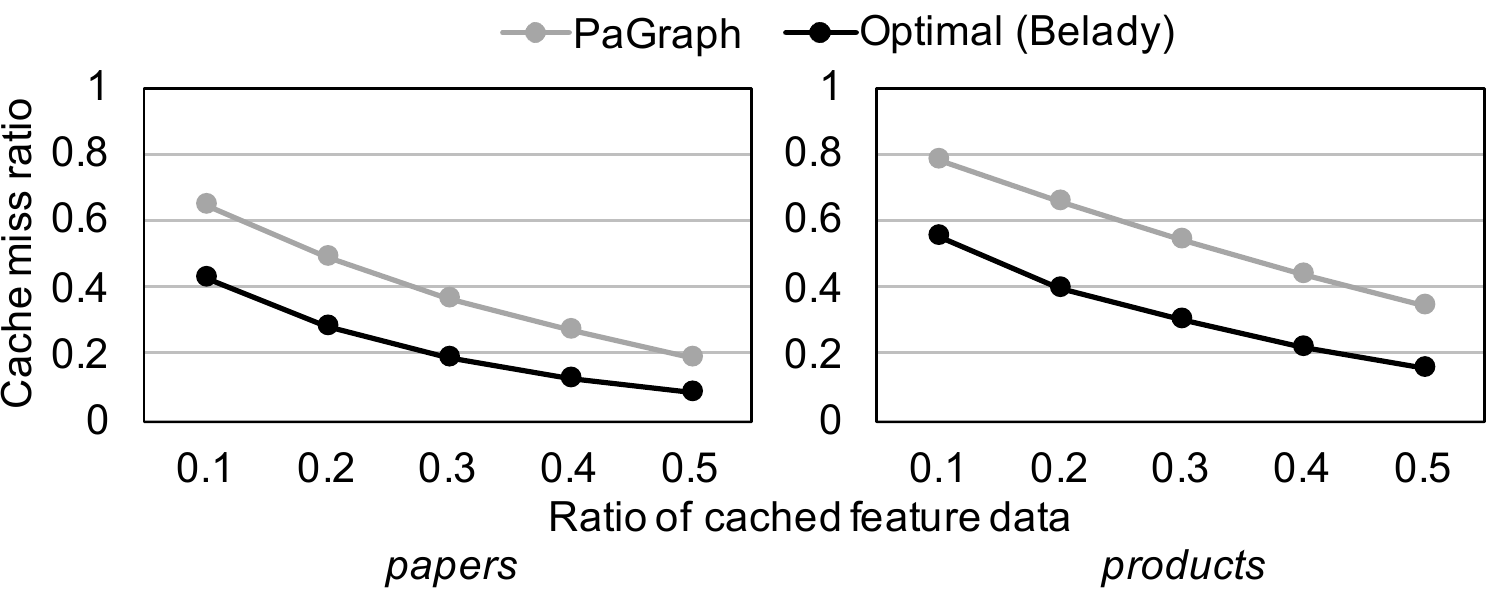}
  \caption{Cache miss ratio of PaGraph and optimal cache with different cache sizes}
  \label{fig:motiv-cache}
\end{figure}
For example, PaGraph~\cite{pagraph}, a state-of-the-art cache design for feature table, caches feature vectors of nodes with a descending order of out-neighbor count. While being moderately effective, this design is sub-optimal as it is static and relies on simple heuristics.  Figure~\ref{fig:motiv-cache} demonstrates cache miss ratio of PaGraph in \texttt{gather} step for the two datasets compared to the \emph{optimal} cache miss ratio with different cache sizes. The optimal cache policy is defined by Belady's cache replacement algorithm, which always evicts data that will not be needed for the longest time in the future~\cite{belady}. The gap from the optimal indicates that there leaves much room for improvement.

\para{Challenge 2. Sub-optimal Pipeline} 
\begin{figure}[t]
  \centering
  \includegraphics[width=\linewidth]{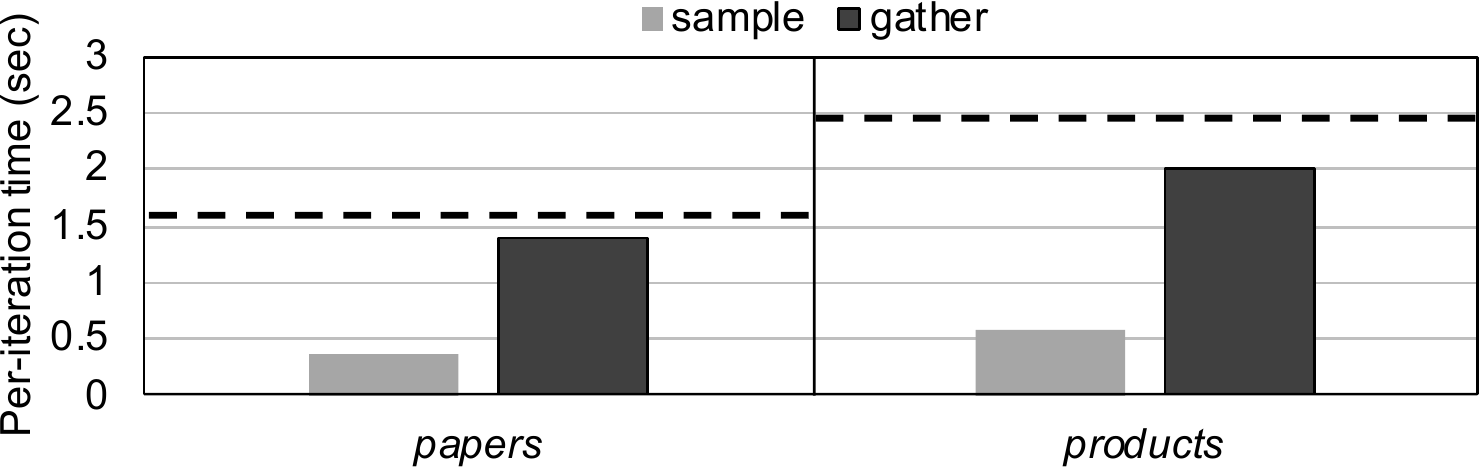}
  \caption{Per-iteration time of stand-alone execution of \texttt{sample} and \texttt{gather} compared to that of pipelined execution of the two operations (represented by dotted line)}
  \label{fig:contention}
\end{figure}
While the two data preparation operations of GNN, \texttt{sample} and \texttt{gather}, are pipelined in the conventional training system, they are not actually an ideal pair for parallel execution especially for disk-based training. This is because they are both I/O intensive operations whose parallel execution can incur resource contention. Under ideal pipelined execution, the time required for data preparation should be determined by the one which takes more time. However, this is not the case as shown in Figure~\ref{fig:contention}. There exists a considerable gap between the stand-alone execution time of the longer operation (\texttt{gather}) and the pipelined execution time (dotted line). This indicates that the conventional pipeline has very limited performance benefit.


\section{\name Design}
\label{sec:design}
\name is a system for efficient training of a very large GNN dataset by using SSD as a memory extension. Specifically, \name targets dataset whose adjacency matrix as well as feature table do not fit in CPU memory. \name reduces I/O traffic from the two I/O intensive operations, \texttt{sample} and \texttt{gather}, by effectively utilizing the main memory space as application-specific in-memory cache. Especially for \texttt{gather}, \name provides the provably optimal caching mechanism. The rearrangement of the training pipeline, inspired by the inspector-executor model~\cite{inex-1,inex-2} in compiler optimization, enables this optimal caching, which would otherwise be infeasible.

The rest of this section is organized as follows. In Section~\ref{sec:design:schedule}, we overview the training pipeline of \name. Section~\ref{sec:design:neighbor-cache} explains the \name neighbor cache (i.e., cache for \texttt{sample}), while Section~\ref{sec:design:feature-cache} and Section~\ref{sec:design:precomputation} explain the \name feature cache (i.e., cache for \texttt{gather}). Section~\ref{sec:design:configuration} provides a guideline for configuring \name's parameters.

\subsection{\name Training Pipeline}
\label{sec:design:schedule}

\para{Overview}
\begin{figure}[t]
  \centering
  \includegraphics[ width=\linewidth]{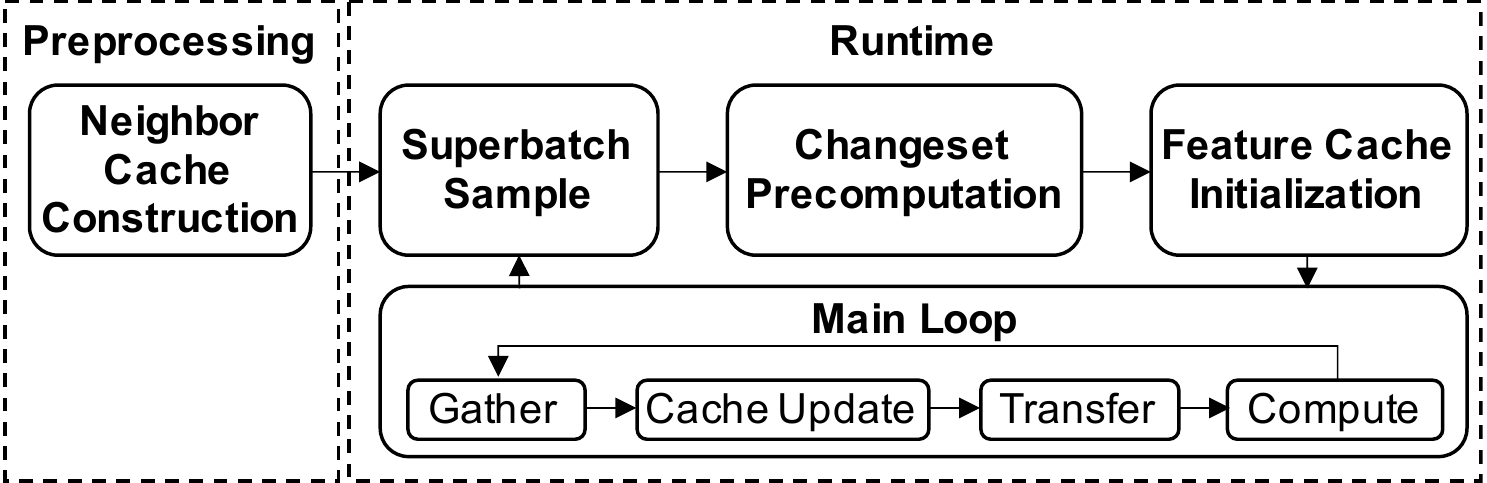}
  \caption{\name training pipeline overview}
  \label{fig:overview}
\end{figure}
Figure~\ref{fig:overview} depicts a high-level overview of \name's training pipeline. After a short preprocessing procedure, \name starts training by iterating the following four stages: \first, \second, \third, and \fourth.  In the \first stage, \name performs sampling for a predefined number of batches, which we call \emph{superbatch}, all at once. With the sampling results, \name finds all the information necessary to manage the feature cache for the following \texttt{gather} operations in the \second stage. Specifically, in this stage, \name determines (i) which feature vectors to prefetch into the feature cache at initialization, and (ii) which feature vectors to insert and which ones to evict from the feature cache (i.e., changeset) at each iteration. After a short transition stage for the \third, \name completes the remaining tasks including \texttt{gather} in the \fourth stage.

\para{Inspector-Executor Execution Model} The inspector-executor
execution model is originally introduced to enable runtime parallelization and scheduling optimization of loops~\cite{inex-1, inex-2}. An inspector procedure runs ahead of the executor to collect information that is available only at runtime, such as data dependencies among array elements. The executor is an optimized version of the original application that utilizes this runtime information to optimize data layout, iteration schedule, and so on. \name embraces this execution paradigm to improve the efficiency of in-memory caching for GNN training. In particular, the first two runtime stages, \first and \second, correspond to the \emph{inspector}, and the \fourth stage to the \emph{executor}. By running ahead the \texttt{sample} operation for the entire superbatch, \name collects complete information about the nodes to be accessed later in the \texttt{gather} stage, thus enabling optimal management of the feature cache.




\begin{figure}[t]
  \centering
  \includegraphics[width=\linewidth]{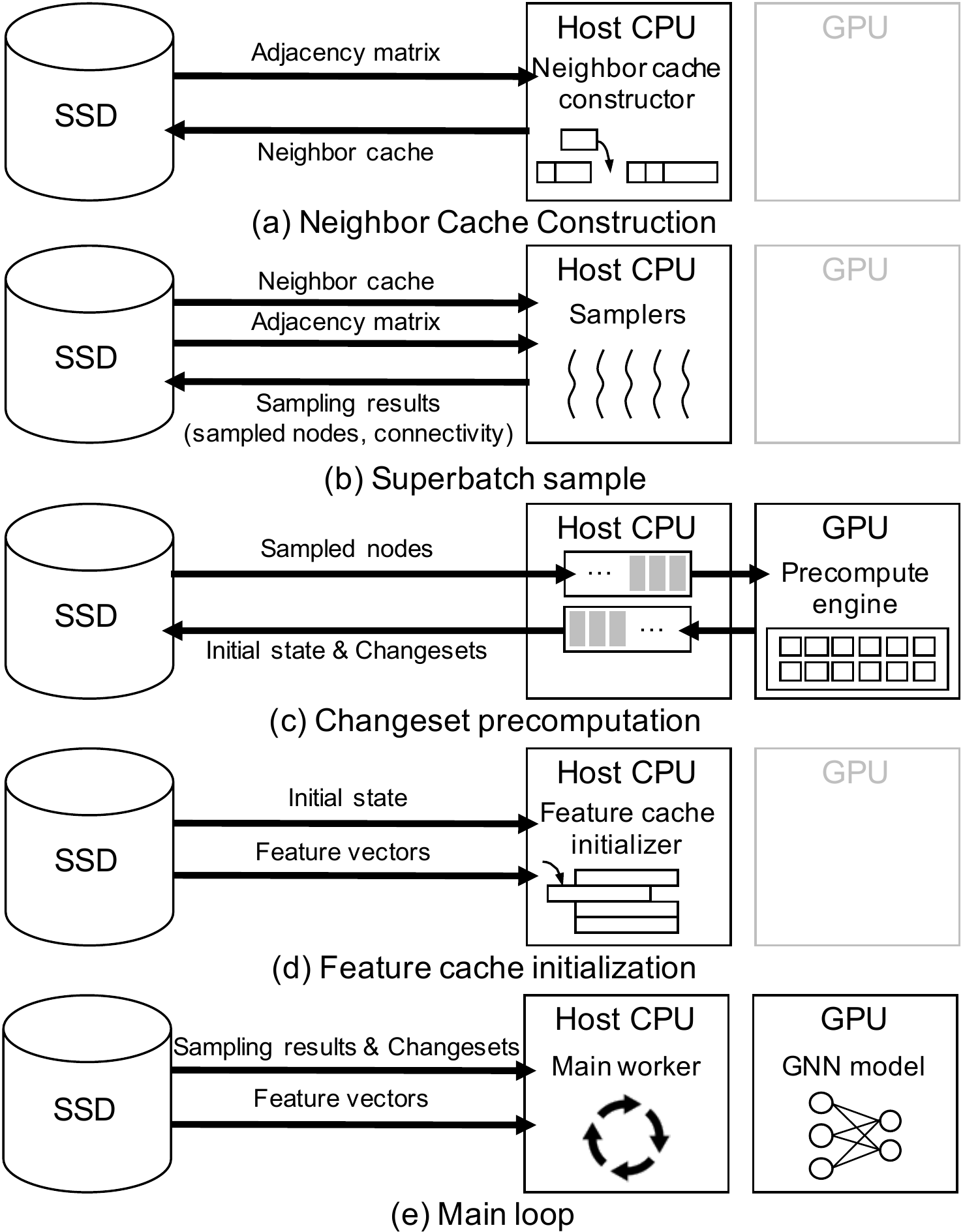}
  \caption{Illustration of \name training pipeline stages}
  \label{fig:schedule}
\end{figure}
\para{Neighbor Cache Construction}
Figure~\ref{fig:schedule}(a) shows this process. Unlike the feature cache which dynamically manages its data, \name uses a static neighbor cache for the whole training process. Therefore, \name constructs the neighbor cache with a given size during offline preprocessing time. To make the neighbor cache, \name examines the graph structure (i.e., adjacency matrix) and picks out \emph{important nodes} whose list of in-neighbors would be cached. The criterion for selecting important nodes will be discussed in Section~\ref{sec:design:neighbor-cache}.
After finishing this construction, \name saves the neighbor cache by dumping it to SSD, which would be loaded at the beginning of each of the following \first stages. This avoids the repeated cost of constructing the neighbor cache, which may include a large number of random reads of which sizes are usually only a few tens or hundreds of bytes.  

\para{Superbatch Sample}
Figure~\ref{fig:schedule}(b) depicts the first stage of \name runtime, \first. In this stage, \name first loads the neighbor cache which has been constructed and stored in SSD during preprocessing. Basically, all memory space except the working buffer for sampling processes can be used for the neighbor cache. After then, multiple sub-processes are launched and sampling is started for as many batches as the superbatch size, $S$.
When accessing the neighbor information during the sampling process, \name first looks up the cache and only reads the data from SSD when it is not present in the cache. The sampling results of a superbatch are then written to SSD. Usually, a sampling for each batch results in two types of data. One is $ids$ which is an 1-D list of all the sampled nodes' IDs. The other is $adj$, a data structure that describes the connectivity among the sampled nodes. \name stores these two data in separate files annotated with the batch index. In total, $2\times S$ files ($ids\_0, ids\_1, ...$ $, ids\_(S-1), adj\_0, adj\_1, ...$ $, adj\_(S-1)$) are generated. The size of each file varies depending on the sampling size, the batch size, and the characteristics of the dataset, but usually ranges from several hundred KB to a few MB. 

\para{Changeset Precomputation}
Figure~\ref{fig:schedule}(c) shows the third stage of \name runtime, \second. Instead of computing a \emph{changeset} (i.e., which features to insert into and evict from the feature cache) every time \name performs \texttt{gather} in \fourth stage, \name precomputes all the changesets beforehand by examining the list of sampled nodes ($ids$ files). This is to accelerate the changeset computation in batch on GPU.
It is difficult to allocate enough memory and computation resources of GPU for the changeset computation in \fourth stage as it involves GPU computation. 
As the total size of $ids$ files may exceed the GPU memory capacity, each $ids$ file is first loaded on the CPU memory and then streamed into GPU when needed.
The results of the changeset precomputation are sent back to CPU, and are stored in SSD also by streaming. Besides the changesets, a list of the feature vectors to prefetch into the cache at initialization is also obtained at this stage. Specifically, $S+1$ files are generated in this step including one for cache initialization ($init$) and the others for cache update for every $S$ iterations ($update\_0, update\_1, ...$ $, update\_(S-1)$). 


\para{Feature Cache Initialization}
Figure~\ref{fig:schedule}(d) shows this process. In this step, \name reads the previously created $init$ file from SSD and constructs the feature cache. This process includes reading feature vectors of the nodes specified in $init$ file as well as building an address table that will be used for cache look-up. 

\para{Main Loop}
Figure~\ref{fig:schedule}(e) illustrates this stage. This stage is where all the remaining GNN training operations for each batch (\texttt{gather}, \texttt{transfer}, and \texttt{compute}) are performed iteratively. In addition, \name performs one more operation, \texttt{cache} \texttt{update}, in between. At each iteration, \name reads a set of $ids$, $adj$ and $update$ files from SSD in the order of the batch index. Then, \name makes batch input by gathering feature vectors according to the $ids$ file from either the cache or SSD, and updates the cache as indicated in the $update$ file. Lastly, the batch input and $adj$ are transferred to GPU in order to perform forward and backward pass as well as model update in the same way as the conventional GNN training system.

\para{Superbatch-level Pipeline}
\begin{figure}[t]
  \centering
  \includegraphics[width=\linewidth]{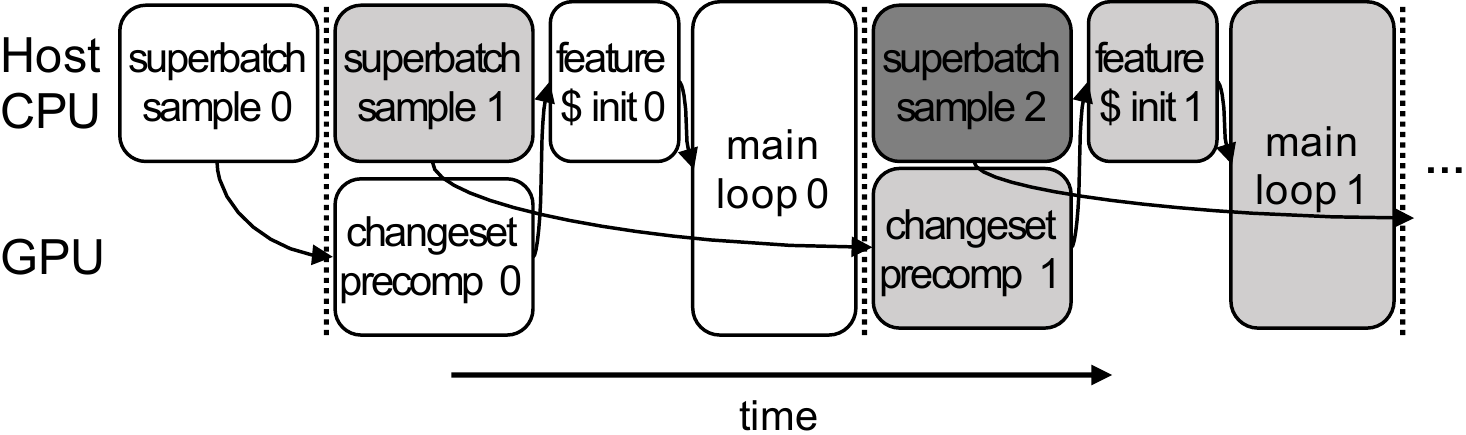}
  \caption{Superbatch-level pipeline of \name}
  \label{fig:ours-pipeline}
\end{figure}
While the four runtime stages for the same superbatch should be serialized, the jobs from different superbatches can be pipelined. Taking this opportunity, \name performs the jobs for different superbatches in a pipelined manner in order to improve end-to-end performance. Specifically, \second for each superbatch is executed in parallel with the \first of the next superbatch.
\first runs on CPU, while \second mainly consumes GPU resources except the I/O overhead of streaming $ids$ files and the changeset precomputation results. This makes these two stages apposite candidates of parallel execution. Figure~\ref{fig:ours-pipeline} visualizes \name's superbatch-level pipeline. While the storage overhead of runtime files is doubled as a result of pipelining, it successfully hides most of the changeset precomputation overhead.

\para{Implications on Training Quality}
The new training schedule of \name has no impact on training quality, as it only changes the execution order of operations which do not have any dependence with each other. It does not require any change in sampling algorithm or GNN model computation.

\subsection{Neighbor Cache}
\label{sec:design:neighbor-cache}
\begin{figure}[t]
  \centering
  \includegraphics[ width=\linewidth]{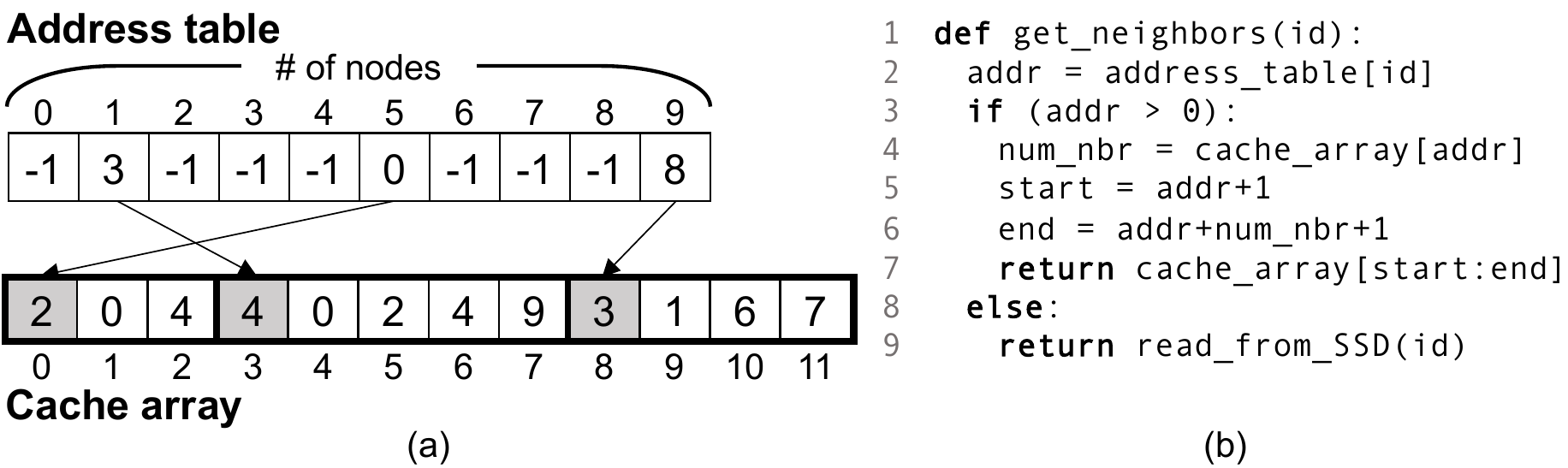}
  \caption{(a) \name neighbor cache structure and (b) pseudocode for its operation}
  \label{fig:neighbor-cache-sturcture}
\end{figure}
\para{Neighbor Cache Policy}
By caching part of the adjacency matrix in memory, \name reduces the number of storage I/Os during sampling. In other words, \name keeps the direct in-neighbors of some important nodes in its neighbor cache in \first stage. To select important nodes, \name adopts a simple metric introduced in Aligraph~\cite{aligraph} which quantifies the trade-off between cost and benefit of caching neighbors of each node in a heuristic way. Specifically, the metric is defined as the ratio between the number of out-neighbors and in-neighbors, as access frequency (benefit) may be roughly proportional to the number of out-neighbors while the cache space overhead (cost) is proportional to the number of in-neighbors. Meanwhile, Aligraph uses the ratio between the number of $k$-hop in-neighbors and out-neighbors, as it caches $k$-hop neighbors ($k\geq2$). However, as \name caches direct neighbors, the ratio between the number of direct in-neighbors and out-neighbors is used as the importance metric for \name. 

\para{Neighbor Cache Structure}
Figure~\ref{fig:neighbor-cache-sturcture}(a) shows the structure of \name's neighbor cache.
Taking a node ID as input, the neighbor cache should return a list of the target node's neighbors if it is cached, or a cache miss signal if not. For this purpose, \name's neighbor cache uses direct addressing. It has an $address\_table$, an 1-D array which has as many elements as the total number of nodes.
Each element of the $address\_table$ contains an index to look up in the $cache\_array$.
A $cache\_array$ is an 1-D array that keeps the neighbor information of the cached nodes for the corresponding node for a cache hit, or an arbitrary negative number (e.g., -1) for a cache miss. Such direct addressing makes its lookup very fast, clearly an $O(1)$ operation, but may incur space overhead owing to the redundancy in its $address\_table$ design. Still, it is affordable as the space overhead of having only a single value for every node is several orders of magnitude smaller than having the whole dataset.

Since each node has a different number of neighbors, the elements of the $cache\_array$ pointed by the $address\_table$ contain the number of neighbors for the corresponding node, and the actual IDs of the neighbors (listed from the next element).
For example, in Figure~\ref{fig:neighbor-cache-sturcture}(a), the neighbor information of Node $1$, $5$ and $9$ are cached since their entries are non-negative.
As $address\_table[1]$ is 3, it means that the neighbor information of Node 1 starts from $cache\_array[3]$.
Since $cache\_array[3]$ is 4, the number of neighbors of Node 1 is 4, and the IDs of its neighbors are listed in 4 consecutive entries, i.e., $cache\_array[4]$ through $cache\_array[7]$.
Figure~\ref{fig:neighbor-cache-sturcture}(b) presents a pseudocode that returns neighbors of a given node using the neighbor cache. Three memory accesses (Line 2, 4, and 7) are needed to fetch the neighbor list from the cache.


\subsection{Feature Cache}
\label{sec:design:feature-cache}
\para{Feature Cache Policy}
The feature cache in \name adopts the provably optimal Belady's algorithm~\cite{belady}. Belady's cache replacement algorithm evicts data with the highest reuse distance at every timestep. Reuse distance for the data is defined as the time until the next access to it. This policy is basically an oracle policy which requires the knowledge of future data accesses, so is infeasible in most cases. In practical settings, only attempts to approximate Belady's cache replacement algorithm have been made~\cite{imitation,backtothefuture,glider}. 

However, it is possible for \name to implement this optimal caching mechanism for feature vectors in the exact form as the sampling results for the whole superbatch are available at the time of \texttt{gather}. Therefore, \name's feature cache dynamically updates its data following Belady's cache replacement algorithm.
Specifically, when updating the cache data at each iteration, the feature cache prioritizes the feature vectors that would be accessed in earlier iterations within the current superbatch.
Initially, to minimize cold misses, \name prefetches the feature vectors that appear at the first few iterations into the feature cache until it gets full.


\para{Feature Cache Structure}
\name's feature cache also uses the same direct addressing as the neighbor cache. Meanwhile, the size of the feature vector is same for all nodes, so the cache array of the feature vector is a simple 2-D array whose row length equals the feature vector size. In the case of a cache hit, the corresponding element in the address table contains the row index of the cache array to look up.
In the case of a cache miss, identical to the case in the neighbor cache, the corresponding element in the address table contains an arbitrary negative value (e.g., -1).


\para{Feature Cache Update}
The changeset precomputation results in three 1-D lists for each iteration which are $in\_ids$, $out\_ids$ and $in\_positions$. $in\_ids$ and $out\_ids$ specifiy the IDs of the nodes to be inserted and evicted, respectively. $in\_positions$ specifies positions of the nodes in $in\_ids$ within a batch input. The batch input refers to the feature vectors collected in a contiguous buffer by \texttt{gather} operation, which is to be transferred to GPU.

\begin{figure*}[t]
  \centering
  \includegraphics[width=\textwidth]{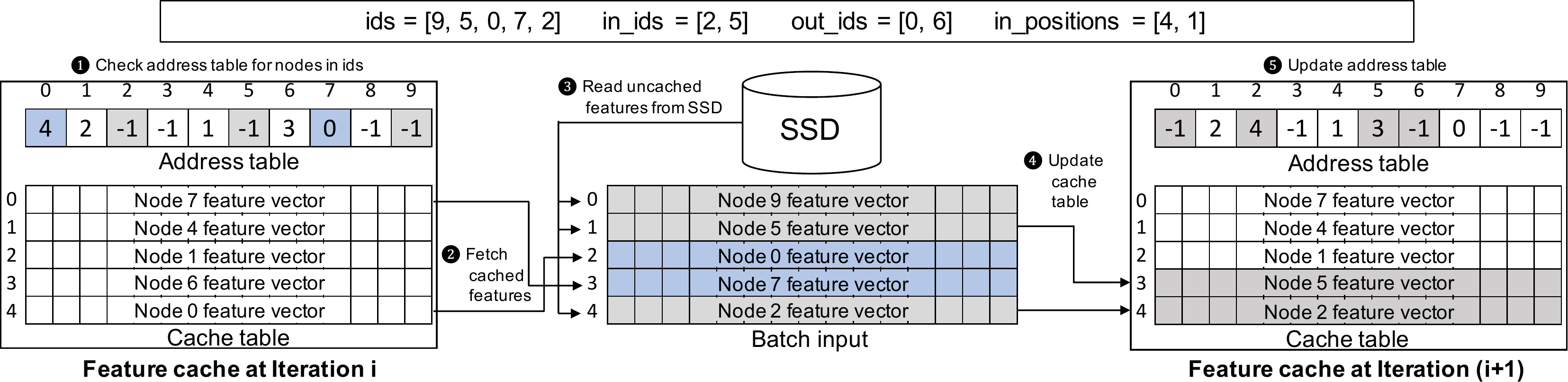}
  \caption{Example of \texttt{cache} \texttt{update} followed by \texttt{gather}}
  \label{fig:iteration}
\end{figure*}

Figure~\ref{fig:iteration} shows the process of \texttt{cache} \texttt{update} followed by \texttt{gather} using an example.
This example shows the change of feature cache when moving on to Iteration ($i+1$) from Iteration $i$.
In this example, the total number of nodes are 10, and five of them are cached (Node $0$, $1$, $4$, $6$, and $7$).
$ids$ refers to the sampled node IDs, while $in\_ids$, $out\_ids$ and $in\_positions$ are cache update information for the current iteration.
For \texttt{gather}, \name first checks $address\_table$ for the nodes in $ids$ \circled{1}.
The feature vectors of the nodes present in the cache (Node $0$ and Node $7$) are fetched from the cache \circled{2}, while the others are read from SSD \circled{3} to make the batch input. Note that the ordering of the feature vectors within the batch input is the same as $ids$.
After the batch input is made, \name performs \texttt{cache} \texttt{update} by referring to $in\_ids$, $out\_ids$ and $in\_positions$.
In the figure, $in\_ids$ and $out\_ids$ indicate that the feature vectors of Node $2$ and Node $5$ should replace the cache lines occupied by the feature vectors of Node $0$ and $6$.
As the feature vectors of Node $2$ and $5$, which are newly loaded from SSD, are buffered in the batch input, \name locates these two vectors with $in\_positions$.
Then, \name puts these vectors into the cache lines originally holding the feature vectors of Node $0$ and Node $6$, whose address can be found by checking $address\_table$ \circled{4}.
Lastly, \name updates $address\_table$ as well \circled{5}, and \texttt{cache} \texttt{update} for Iteration $i$ ends.

\subsection{Changeset Precomputation Algorithm}
\label{sec:design:precomputation}

\para{Problem Formulation}
Computing the changeset is no more than simulating the cache state (i.e., node IDs present in the cache) of every iteration as the changeset can be obtained by comparing the difference between the two consecutive cache states. In other words, it is to solve the following recurrence relation for $i$ from $0$ to $S-2$, where $S$ is the superbatch size.
\begin{equation*}
C_{i+1} = f_{Belady}(union(C_{i},ids\_i))
\end{equation*}
where $C_{i}$ is the cache state at Iteration $i$.
The initial cache state ($C_{0}$) and the lists of node IDs that would be accessed at each iteration of superbatch ($ids\_0, ids\_1, ..., ids\_(S-2)$) are given as inputs.
$f_{Belady}$ is a function that selects the elements in the given list, prioritizing those that are to be accessed earlier than others, as many as the number of cache entries.
As all the feature vectors of the nodes in $ids\_i$ are loaded onto memory at Iteration $i$, the \textit{union} of $C_{i}$ and $ids\_i$ are potential candidates to be included in $C_({i+1)}$.

The main challenge in solving $f_{Belady}$ is to find out the \emph{next accessed iterations} of the given nodes in $union(C_{i}, ids\_i$) at each iteration.
Here, we denote the iteration that the node is to be accessed next by the node's \emph{next accessed iteration}.
A straightforward approach is to examine all the future data access traces ($ids\_({i+1})$, $ids\_({i+2})$, ...) in order to check in which iteration the feature vector of each node in $union(C_{i},ids\_i)$ would be accessed next time. Such a na\"ive approach, however, results in the worst case complexity of $O(S^2)$ whose cost can be substantial as we increase the superbatch size. To prevent the changeset precomputation stage from being the major bottleneck, we propose an efficient algorithm with just an $O(S)$ complexity.

\para{Sketch of Our Algorithm}
\name solves the problem with only three passes over the access traces. With the first two passes, \name builds a data structure specially designed for the incremental tracking of each node's next accessed iteration. With this data structure, in the last pass, \name simulates a whole sequence of cache states. The following paragraphs discuss how this data structure looks like and operates as well as how the end-to-end algorithm works in details.

\begin{figure*}[t]
  \centering
  \includegraphics[width=\textwidth]{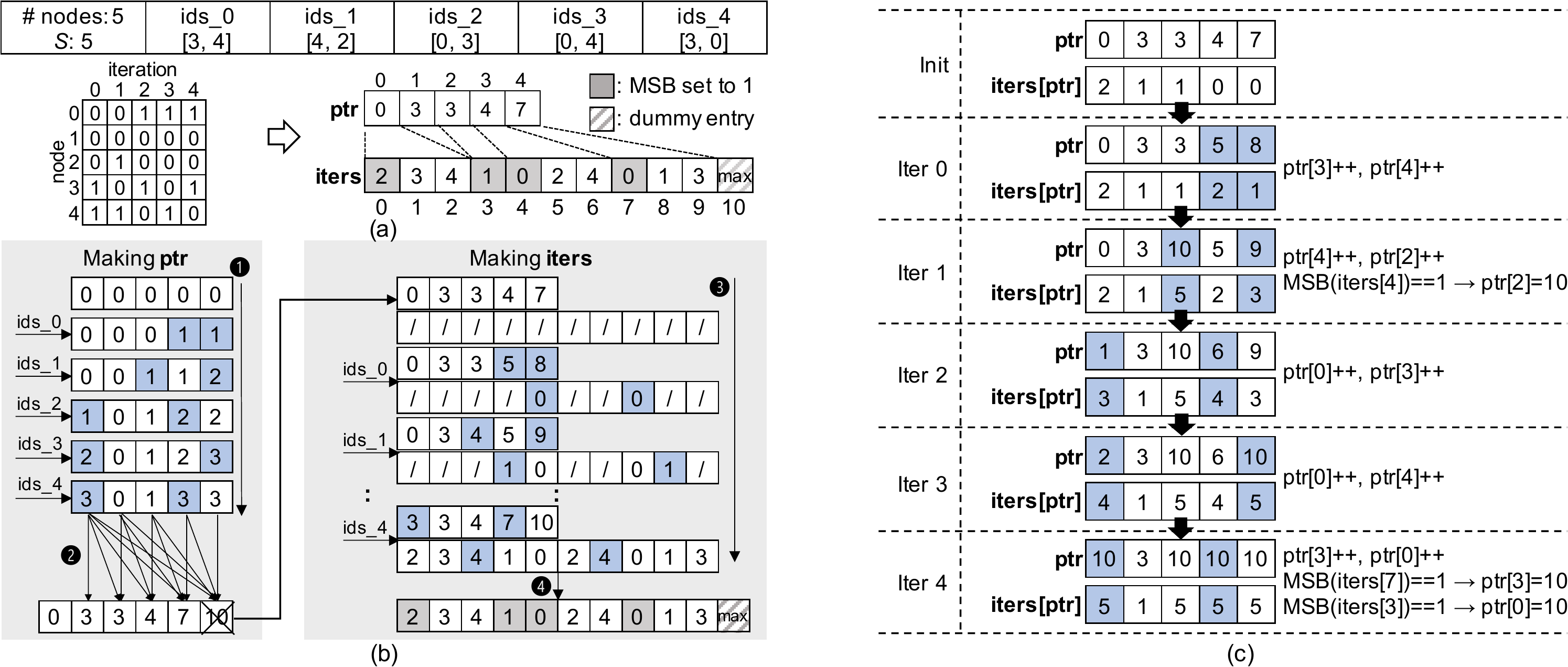}
  \caption{Example for next accessed iteration tracking of \name. (a) shows data structures for the tracking while (b) shows the process of making it. (c) shows how \name tracks the next accessed iteration.}
  \label{fig:tracking}
\end{figure*}

\begin{figure}[!t]
\begin{minipage}[b]{1.0\hsize}
\begin{lstlisting}[style=myStyle]
# num_entries: the number of cache entries
# ids_list: the list of data access traces ([ids_0, ids_1, ..., ids_S-1]) stored in SSD

current_cache_state = initial_cache_state
for ids in ids_list:
  incoming_ids = Load(ids)
  
  ptr[incoming_ids] += 1
  for id in incoming_ids where MSB(iters[ptr[id]]) is set:
    ptr[id] = len(iters)-1

  candidates = Union(current_cache_state, incoming_ids)
  next_access_iters = iters[ptr[candidates]]
  
  # select N elements in candidates with the smallest next accessed iteration
  new_cache_state = TopKSmallest(candidates, next_access_iters, num_entries)
  
  # find relative complement of new_cache_state & current_cache_state with respect to each other  
  in_ids = new_cache_state - current_cache_state
  out_ids = current_cache_state - new_cache_state
  in_positions = IndexOf(in_ids, incoming_ids)
  
  Save((in_ids, out_ids, in_positions))
 
\end{lstlisting}
\end{minipage}
\caption{Pseudocode for end-to-end changeset precomputation process}
\label{fig:precomputation-pseudocode}
\end{figure}

\para{Tracking Next Accessed Iteration}
Figure~\ref{fig:tracking}(a) shows the data structure that \name utilizes for tracking next accessed iterations of nodes with a simple example, where both the number of nodes and the superbatch size are five. We assume that two nodes are sampled for each iteration, which means that the length of each $ids$ is two. 
Conceptually, the access traces ($ids\_0, ids\_1, ..., ids\_4$) can be represented as a binary matrix where each row corresponds to each node while each column corresponds to each iteration.
($i$,$j$) is set to $1$ if Node $i$ appears at Iteration $j$, or $0$ otherwise. 
What \name maintains for next accessed iteration tracking is a sort of CSR-format of this matrix which consists of two arrays: \textit{iters} and \textit{ptr}. 
\textit{iters} lists accessed iterations of each node in sequence from Node $0$.
Each node's accessed iterations are kept in sorted order. \textit{ptr} array has elements as many as the number of nodes each of which points to the start of the corresponding node's accessed iterations in \textit{iters}. For example, \textit{ptr}[$3$] is $4$, which means that from \textit{iters}[$4$], the iterations in which Node $3$ appears ($0$, $2$, $4$) are listed in order.
The most significant bit (MSB) of the elements in \textit{iters} pointed by \textit{ptr} is set to $1$ and a dummy entry whose value is set to the maximum value for a given datatype is appended at the end of \textit{iters}.  

The construction of this data structure requires two passes over the access traces. Figure~\ref{fig:tracking}(b) shows this process. In the first pass, \name counts the number of appearances of each node \circled{1}. Then, \name performs cumulative sum over the count results and completes making \textit{ptr} by inserting zero value in the front and deleting the last element \circled{2}. In the second pass, \name makes \textit{iters} with a guide of \textit{ptr} \circled{3}.
At Iteration $0$, Node $3$ and Node $4$ are accessed.
So \name sets 0 for the elements pointed by \textit{ptr}[$3$] ($4$) and \textit{ptr}[$4$] ($7$), then increments \textit{ptr}[$3$] and \textit{ptr}[$4$] by one.
This process is repeated for the remaining iterations.
At last, \name restores \textit{ptr} by a simple shift and completes \textit{iters}
construction by appending a dummy entry and setting MSBs for elements pointed by \textit{ptr} to 1 \circled{4}.

Using these two arrays, \name can track next accessed iteration of the nodes with a single additional pass over the access traces, which completes the cache state simulation. \name updates \textit{ptr} for each iteration in a way that it can always find out the next accessed iterations of all nodes by looking at the elements in \textit{iters} pointed by \textit{ptr}. Figure~\ref{fig:tracking}(c) shows this process with the same example as before. In the figure, \textit{ptr} and a subset of of \textit{iters} pointed by \textit{ptr} (\textit{iters}[\textit{ptr}]), which would indicate the next accessed iteration of the nodes, are shown at initialization and each iteration. Note that \name does not maintain \textit{iters}[\textit{ptr}] in fact but only looks up a necessary part on demand.
We show the change of \textit{iters}[\textit{ptr}] along with \textit{ptr} just for demonstration. At initialization, \textit{iters}[\textit{ptr}] denotes the first accessed iteration of the nodes. When simulating each iteration, \textit{ptr} values for the accessed nodes are incremented by $1$.
This is natural as the next accessed iteration of the currently accessed nodes should be updated. For example. at Iteration $0$ in the figure, \textit{ptr}[$3$] and \textit{ptr}[$4$] are incremented by $1$.
If the MSB of the \textit{iters} element pointed by the updated \textit{ptr} value is $1$, however, that \textit{ptr} value is set to point the last element of \textit{iters}, which is the dummy entry.
This is to regard the next accessed iteration of such a node which would not appear again in the future as a very large value.
For example, at Iteration $1$ in the figure, \textit{ptr}[$2$] is set to $10$, the index of the dummy entry in \textit{iters}, since the MSB of \textit{iters}[$4$] is set to $1$. The same process is repeated until the end of the simulation. 

\para{End-to-end Algorithm}
Figure~\ref{fig:precomputation-pseudocode} is a pseudocode that shows the end-to-end process of the changeset precomputation which eventually produces the three lists ($in\_ids$, $out\_ids$, and $in\_positions$) for each iteration. For each iteration, the data access trace for each iteration (\textit{incoming\_ids}) originally stored in SSD is loaded into GPU in order (Line 6). The \textit{ptr} is updated to remain up-to-date (Line 8-10). With the updated \textit{ptr}, the next accessed iterations are obtained for the candidates which is the union between the IDs in the current cache and incoming node IDs (Line 12-13).
After that, the new cache state is derived by selecting top-\texttt{num\_entries} smallest among the next accessed iterations of the candidates (Line 16). By computing the set difference between the new cache state and the current cache state, $in\_ids$ and $out\_ids$ can be obtained (Line 19-20), after which $in\_positions$ is computed by figuring out the positions of each node in $in\_ids$ within \textit{incoming\_ids}. Finally, the three resulting lists are sent back to the host and saved in SSD.

\subsection{Configuring \name}
\label{sec:design:configuration}
\para{Superbatch Size}
For performance, it is \textit{always} better to increase the superbatch size. This is because a large superbatch can amortize the cost of switching between \first and \fourth, which includes loading the neighbor cache and initializing the feature cache. This switch cost depends on the cache size and remains constant with respect to the superbatch size. However, it is not possible to increase the superbatch size indefinitely for the following reasons. First, GPU memory size imposes a hard limit on the superbatch size. The amount of GPU memory required to perform the changeset precomputation increases with the superbatch size. However, this is not often the case unless a very large superbatch size (e.g., $10000$) is used. Instead, the storage overhead of runtime files may be a practical factor that limits the superbatch size. One can increase the superbatch size as much as the storage constraint permits. In our evaluation setting, 100GB of runtime files are a good trade-off between the performance and storage overhead (Section~\ref{sec:eval:sb-size}). Since the runtime file size per iteration is relatively constant throughout the training process, one can readily approximate the storage overhead via a short offline profiling.



\para{Cache Size}
The size of the neighbor cache and the feature cache should be determined before the beginning of the training. Since the peak memory usage at each iteration of \first and \fourth remains stable throughout the training process, we can again determine the cache size by a short offline profiling. We set the cache size by subtracting the peak memory usage at each stage (\first for the neighbor cache and \fourth for the feature cache), and the size of additionally reserved region for memory fluctuations from the total memory size.


\section{Implementation}
\label{sec:impl}
We have implemented \name by extending PyTorch Geometric (PyG)~\cite{pyg}, a popular open-source framework for GNN training built upon PyTorch~\cite{pytorch}, as follows. First, we have created new extensions for sampling and gathering to support the user-level in-memory cache. Specifically for sampling, we have modified PyTorch Sparse C++ backend which PyG calls for sampling-related operations. We use \texttt{pread} syscall with \texttt{O\_DIRECT} flag to bypass OS page cache when accessing to graph data stored in SSD. Second, we have made an extension also for cache updates. To copy sparsely located data in a tensor to the specified locations of another tensor, PyTorch tensor indexing API requires to collect the data from the source tensor in a contiguous buffer first and then scatter them to the destination tensor, which incurs redundant \texttt{memcpy}. Our extension eliminates such waste by directly copying data from the source to the destination. This process is parallelized by OpenMP. Third, we have modified \texttt{NeighborSampler} class of PyG in a way that it stores the sampling results to disk in order instead of putting them into a shared queue. To load the saved files at each iteration of the \fourth stage, we use a custom lightweight multi-threaded dataloader. Lastly, we take advantage of highly-optimized CUDA kernels of PyTorch for changeset precomputation.

\section{Evaluation}
\label{sec:eval}

\subsection{Methodology}
\label{sec:eval:setup}
\para{System Configurations}
Table~\ref{tbl:sysconfig} summarizes our system configurations. We evaluate \name on a Gigabyte R281-3C2 server with an 8-core CPU (16 logical cores with hyper-threading), an NVIDIA V100 GPU, and a Samsung PM1725B NVMe SSD.

\begin{table}[!t]
\centering
\caption{System configurations} \label{tbl:sysconfig}
\begin{adjustbox}{width=\linewidth}
\begin{tabular}{|l|lll|}
\hline
CPU     & \multicolumn{3}{l|}{Intel Xeon Gold 6244 CPU 8-core @ 3.60 GHz}                                                                               \\ \hline
GPU     & \multicolumn{3}{l|}{NVIDIA Tesla V100 16GB PCIe}                                                                                              \\ \hline
Memory  & \multicolumn{3}{l|}{Samsung DDR4-2666 64GB (32GB $\times{}$ 2)}                                                                               \\ \hline
Storage & \multicolumn{3}{l|}{\begin{tabular}[c]{@{}l@{}}Samsung PM1725b 8TB PCIe Gen3 8-lane\end{tabular}} \\ \hline
S/W & \multicolumn{3}{l|}{\begin{tabular}[c]{@{}l@{}}Ubuntu 18.04.5 \& CUDA 11.4 \& Python 3.6.9 \& PyTorch 1.9 \end{tabular}} \\ \hline
\end{tabular}
\end{adjustbox}
\end{table}

\para{Model and Dataset}
\rev{We use 3-layer GraphSAGE~\cite{graphsage} and 2-layer GCN~\cite{node-classification} for  evaluation. Both models have a hidden dimension of 256. For GraphSage, we set a sampling size to (10,10,10). By default, we set batch size to 1000.} To evaluate \name on a billion-scale graph, we scale four real-world datasets: \textit{ogbn-papers100M} (\textit{papers})~\cite{ogb}, \textit{ogbn-products} (\textit{products})~\cite{ogb}, \textit{com-friendster} (\textit{Friendster})~\cite{snap}, and \textit{twitter-2010} (\textit{Twitter})~\cite{snap} following the methodology in ~\cite{gnnssd}. Specifically, we use a graph expansion technique~\cite{scalable}, which adapts Kronecker graph theory~\cite{kronecker} to preserve innate distributions of recipe graphs like power-law degree distribution and community structure. We then randomly choose 10\% of the nodes to serve as a training set. 
\rev{Even with this split of the dataset, the working set may contain
the feature vectors of the whole dataset as GNN training takes the feature vectors of not only the nodes in the training set but also their $k$-hop neighbors as input.}
By default, we set the feature dimension of all datasets to 256. The whole datasets are assumed to be stored in SSD during training except the pointer array in a CSC-formatted adjacency matrix. While the pointer array takes only about a few GBs, it is very frequently accessed during the \texttt{sample} stage. Thus, we keep it in memory when performing \texttt{sample}. Table~\ref{tbl:datasets} summarizes the key features of the datasets.

\begin{table}[t]
\centering
\caption{Graph datasets} \label{tbl:datasets}
\begin{adjustbox}{width=\linewidth}
\begin{tabular}{|c|cc|ccc|}
\hline
                & \multicolumn{2}{c|}{Original}          & \multicolumn{3}{c|}{Large-scale}                                                                                \\ \hline
Dataset         & \multicolumn{1}{c}{nodes} & edges & \multicolumn{1}{c}{nodes} & \multicolumn{1}{c}{edges} &  size \\ \hline
\textit{ogbn-papers100M} & \multicolumn{1}{c}{111.06M}        & 1.62B       & \multicolumn{1}{c}{444.24M}        & \multicolumn{1}{c}{14.24B}      &  569GB      \\ \hline
\textit{ogbn-products}   & \multicolumn{1}{c}{2.45M}            & 61.86M             & \multicolumn{1}{c}{220.41M}            & \multicolumn{1}{c}{20.24B}                  & 388GB           \\ \hline
\textit{com-friendster}  & \multicolumn{1}{c}{65.61M}            & 1.81B             & \multicolumn{1}{c}{262.43M}            & \multicolumn{1}{c}{15.48B}                   & 393GB           \\ \hline
\textit{twitter-2010}    & \multicolumn{1}{c}{41.65M}            & 1.47B             & \multicolumn{1}{c}{208.26M}            & \multicolumn{1}{c}{14.05B}                 & 326GB           \\ \hline
\end{tabular}
\end{adjustbox}
\end{table}

\para{Comparison Baselines} We compare \name with two baselines: PyG+, Ali+PG. PyG+ refers to the PyG framework modified to support disk-based GNN training over a memory-mapped graph dataset. It follows the conventional GNN training pipeline explained in Section~\ref{sec:back:pipeline}. We used \texttt{memmap} function of NumPy~\cite{numpy} to create a memory-map to an array stored in a binary file and converted the memory-mapped array into a PyTorch tensor.
As an optimization for PyG+, we disable the readahead feature, which only degrades the performance of GNN training as it incurs a large number of random disk accesses.
In addition, we set the number of threads to use for I/O intensive operations like \texttt{gather}, more than twice the number of cores to better utilize the disk bandwidth instead of using the default PyTorch setting which is the number of physical cores. Ali+PG is PyG+ augmented by an application-specific in-memory caching mechanism for \texttt{sample} and \texttt{gather}. Specifically, an Aligraph-style cache~\cite{aligraph} is used for the neighbor cache and a PaGraph-style~\cite{pagraph} cache is used for the feature cache.  
For Ali+PG, we tune the sizing of the two caches by splitting the available memory space on the ratio of 0:100, 25:75, 50:50, 75:25, and 100:0 for the neighbor cache and the feature cache, and report the one that yields the best performance. Ali+PG represents a strong baseline by integrating two state-of-the-art caching techniques. For all schemes, we overlap the CUDA operations (\texttt{transfer}, \texttt{compute}) on GPU with the CPU operations to reduce data stalls. 
\rev{For each data point we report the mean of three measurements with an error bar.}

\subsection{Overall Performance}
\label{sec:eval:perf}

\begin{figure}[t]
  \centering
  \includegraphics[width=\linewidth]{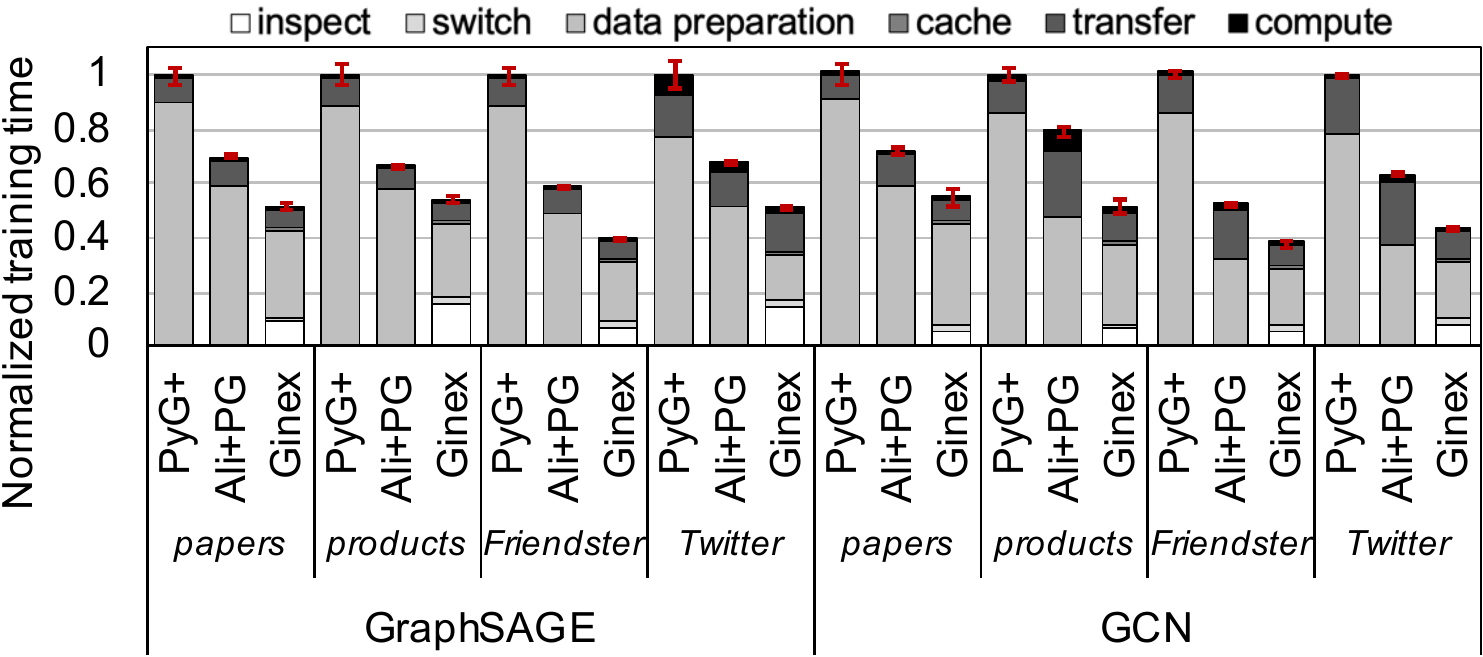}
  \caption{\rev{Normalized training time breakdown of PyG+, Ali+PG, and \name. Smaller is better.}}
  \label{fig:eval:main}
\end{figure}

\begin{figure}[t]
  \centering
  \includegraphics[width=\linewidth]{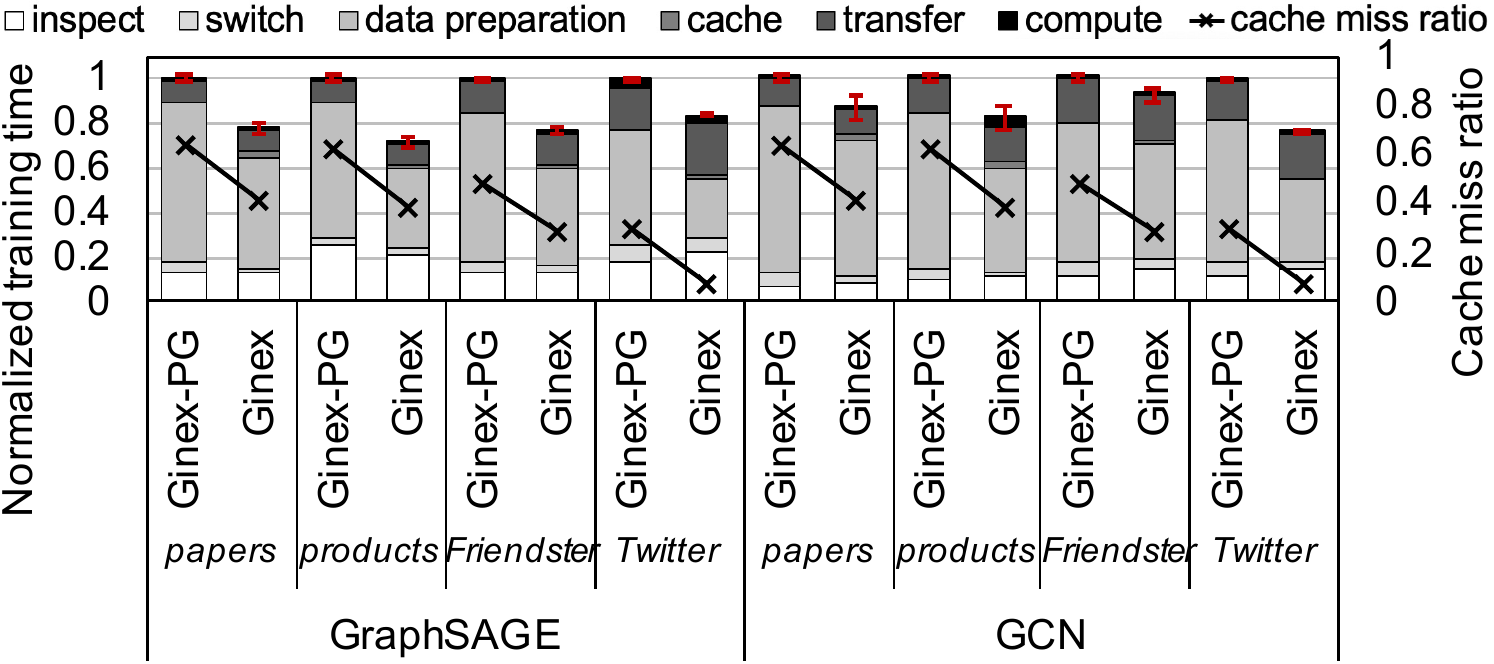}
  \caption{\rev{Normalized training time breakdown of \name-PG and \name}}
  \label{fig:eval:opt}
\end{figure}

We first measure and break down the training time of PyG+, Ali+PG and \name for the four datasets. For GraphSAGE, the superbatch size of \name is set to 3300, 2100, 3600, and 6400 for \textit{papers}, \textit{products}, \textit{Friendster}, and \textit{Twitter}. For GCN, we set the superbatch size of \name to 2500, 300, 900, and 900, respectively. We derive these values via the offline profiling-based heuristic in Section~\ref{sec:design:configuration}. The actual size of runtime files falls very close to our target (100 GB) within 3\%.


Figure~\ref{fig:eval:main} shows the results. The training time of PyG+ and Ali+PG can be broken down into three components: data preparation, transfer, and compute. Data preparation time refers to the amount of time that the CUDA operations are stalled either by \texttt{sample} or \texttt{gather}. Meanwhile, the training time of \name has three more components: inspect, switch, and cache update. Inspect time refers to the time consumed for the pipelined execution of superbatch sample and changeset precomputation. In all cases the time for changeset precomputation completely hidden by superbatch sample, and the inspect time is equal to the time for superbatch sample. Switch time is the time spent for initializing the feature cache. For \name, data preparation time includes not the time for \texttt{sample} but that for \texttt{gather} and runtime file loading.

For all workloads \name demonstrates the superior performance. For GraphSAGE the speedups of \name range 1.86-2.50$\times$ over PyG+ and 1.23-1.47$\times$ over Ali+PG. For GCN \name is 1.83-2.67$\times$ and 1.28-1.57$\times$ faster than PyG+ and Ali+PG, respectively. This performance gains are attributed to the optimal caching scheme for \texttt{gather} outweighing the cost for it. The overhead of the serialization of \texttt{sample} and \texttt{gather} is shown to be insignificant as expected. The switch time and cache update time also account for a minor portion (less than 10\%). All these benefits come with a moderate storage cost of about 145 GB (for runtime files and neighbor cache), which is less than a half of the smallest dataset being used. 

\subsection{Impact of Optimal Feature Cache}
\label{sec:eval:sb-size}
In Figure~\ref{fig:eval:opt}, we evaluate \name with its cache policy replaced by PaGraph (\name-PG) to measure the impact of \name's optimal feature cache in isolation. We report the normalized training time and the cache miss ratio. \name consistently demonstrates significantly lower miss ratio, which leads to a proportional reduction in data preparation time. The adoption of the optimal feature cache, however, results in a slightly longer inspect time. While the changeset precomputation can be hidden under the superbatch sample, it may slow down superbatch sample as it involves disk I/Os to stream input and output of the changeset precomputation. However, the increase of inspect time is limited to less than 20\%, which is small compared to the data preparation time reduction.

\subsection{Sensitivity Study}
\label{sec:eval:sensitivity}
We conduct a sensitivity study with varying four parameters that may affect performance: superbatch size, feature dimension, memory size, and batch size. We only use GraphSAGE for this study.

\para{Impact of Superbatch size}
\begin{figure}[t]
  \centering
  \includegraphics[width=\linewidth]{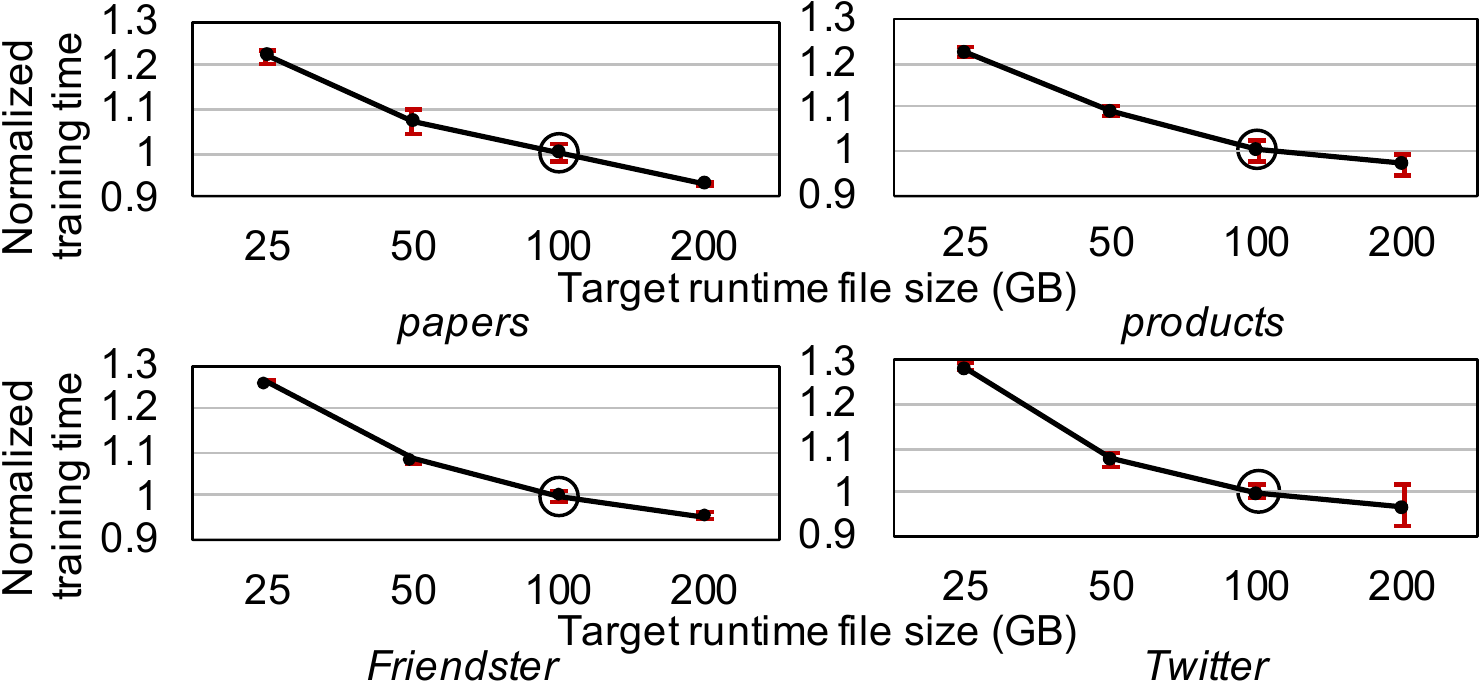}
  \caption{Training time of \name normalized to the default superbatch size (circled) on varying superbatch sizes}
  \label{fig:eval:sbsize}
\end{figure}
In Figure~\ref{fig:eval:sbsize}, we evaluate \name with different superbatch sizes to quantify its performance impact. Specifically, we adjust the target runtime file size from 25 GB to 200 GB and report the training time normalized to that of the default setting in which the target runtime file size is 100 GB. Generally, the larger the superbatch size is, the better the performance tends to be. This is mainly because the switch time, which remains constant, is amortized. However, increasing the superbatch size eventually gives diminishing returns beyond a certain point. For example, doubling the runtime file size from 100 GB to 200 GB results in only a marginal decrease of the training time ($7.22\%$, $3.15\%$, $4.81\%$, and $3.16\%$ for the four datasets). Thus, it is an effective heuristic to set the target runtime file size (100GB in our setting) based on the storage capacity constraint as discussed in Section~\ref{sec:design:configuration}.

\para{Impact of Feature Dimension}
\begin{figure}[t]
  \centering
  \includegraphics[width=\linewidth]{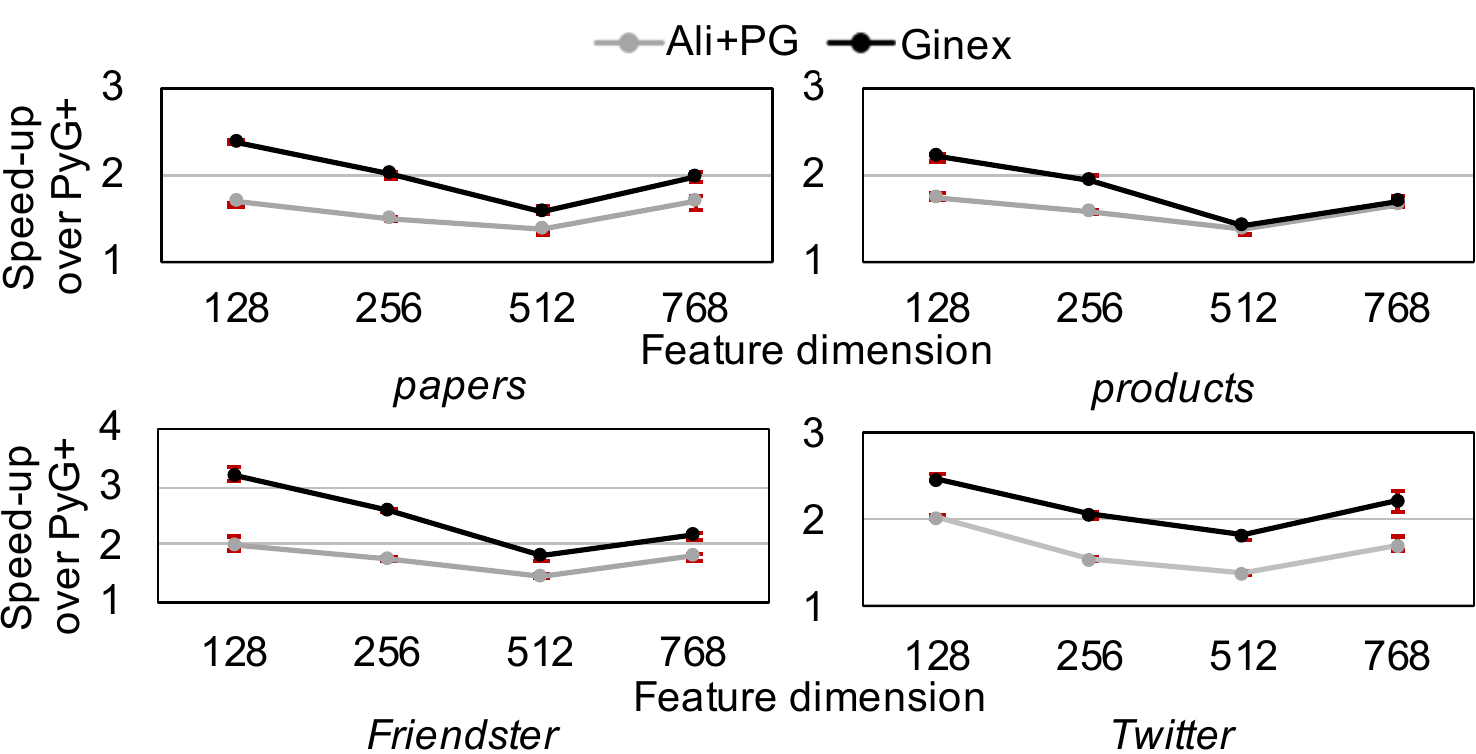}
  \caption{Speedup of Ali+PG and \name over PyG+ with varying feature dimensions}
  \label{fig:eval:feature}
\end{figure}
In Figure~\ref{fig:eval:feature}, we evaluate the impact of feature dimension on performance. Specifically, we run experiments scaling the feature dimension to $\times0.5$, $\times2$ and $\times3$ of the default setting (256). We report the speedup of \name and Ali+PG over PyG+. While \name is consistently faster than PyG+ as well as Ali+PG, the degree of its improvement varies with the feature dimension. For the feature dimension of 128, 256 and 512, the higher the feature dimension, the smaller the gap among the three schemes. This can be explained by two factors.
First, the relative miss penalty of in-memory cache gets smaller as the feature dimension increases.
Regardless of whether the feature dimension is 128, 256, or 512, the miss penalty is the same as a whole 4KB page should be read from SSD. On the other hand, the hit time linearly increases with the feature dimension. Thus, the reduction of the number of cache misses is more critical when the feature dimension is low, which makes the impact of \name's optimal caching more significant.
Second, the relative cache size compared to the whole feature table size gets smaller with higher feature dimension.
For example, when using a feature dimension of 512 in \textit{papers} dataset, only about 5\% of the whole feature data can be kept in \name's feature cache. In such case, there is usually no significant difference in terms of cache miss ratio between different mechanisms. 

Meanwhile, the experimental results with the feature dimension of 768 do not follow the trend discussed above. The gap among the three schemes becomes slightly greater.
This is because of the alignment issue. When the feature dimension is 768, the size of each feature vector is 3KB assuming 32-bit floating point format. In this case, a single feature vector may often span two 4KB pages, which means that one should read 8KB in total from SSD to access it. This does not happen when the feature dimension is 128, 256 and 512 as the page size is aligned with the feature vector size. This issue amplifies the miss penalty, thereby making the impact of in-memory caching mechanism more pronounced.
 
\para{Impact of Memory Size}
\begin{figure}[t]
  \centering
  \includegraphics[width=\linewidth]{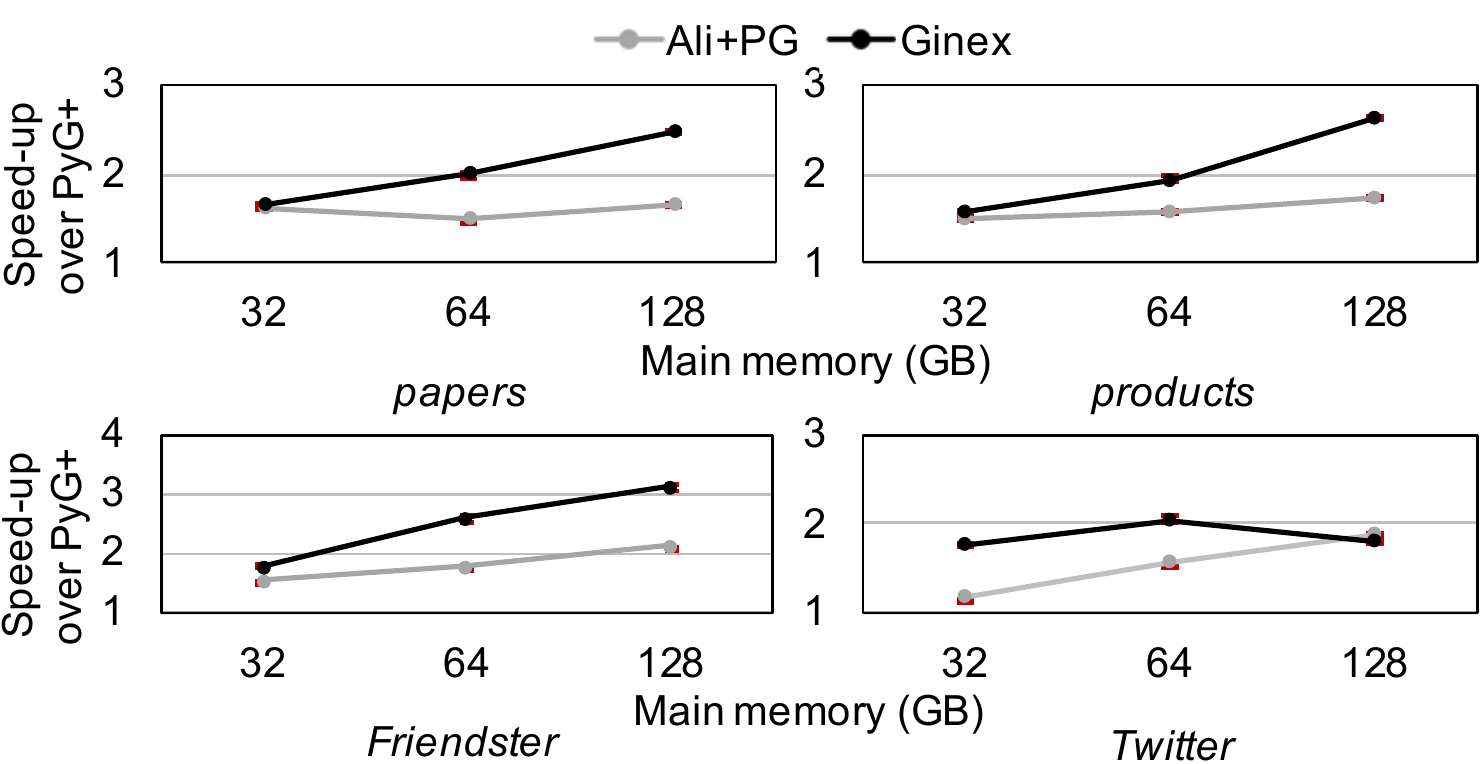}
  \caption{\rev{Speedup of Ali+PG and \name over PyG+ with varying main memory size}}
  \label{fig:eval:cache-size}
\end{figure}
Figure~\ref{fig:eval:cache-size} shows the impact of memory size. We scale the memory size to 0.5$\times$ and 2$\times$ of the default size (64 GB). The performance gap between the two caching schemes tends to get narrower as the memory size gets smaller. It is because only about 10 GB of memory can be used for the cache after sparing enough workspace at 32 GB. Too small cache leads to equally poor caching performance for both schemes. Meanwhile, the \textit{Twitter} dataset demonstrates different trend. Due to the high inter-batch locality, \name can achieve substantial performance gains even at 32 GB. However, both schemes are comparable at 128 GB. This is because the caching performance of \name already gets saturated with a very low cache miss ratio (<8\%) at 64 GB (see Figure~\ref{fig:eval:opt}). Thus, allocation of an additional memory space gives very little benefits to \name, whereas Ali+PG gets a substantial boost from it.



\para{\rev{Impact of Batch Size}}
\begin{figure}[t]
  \centering
  \includegraphics[width=\linewidth]{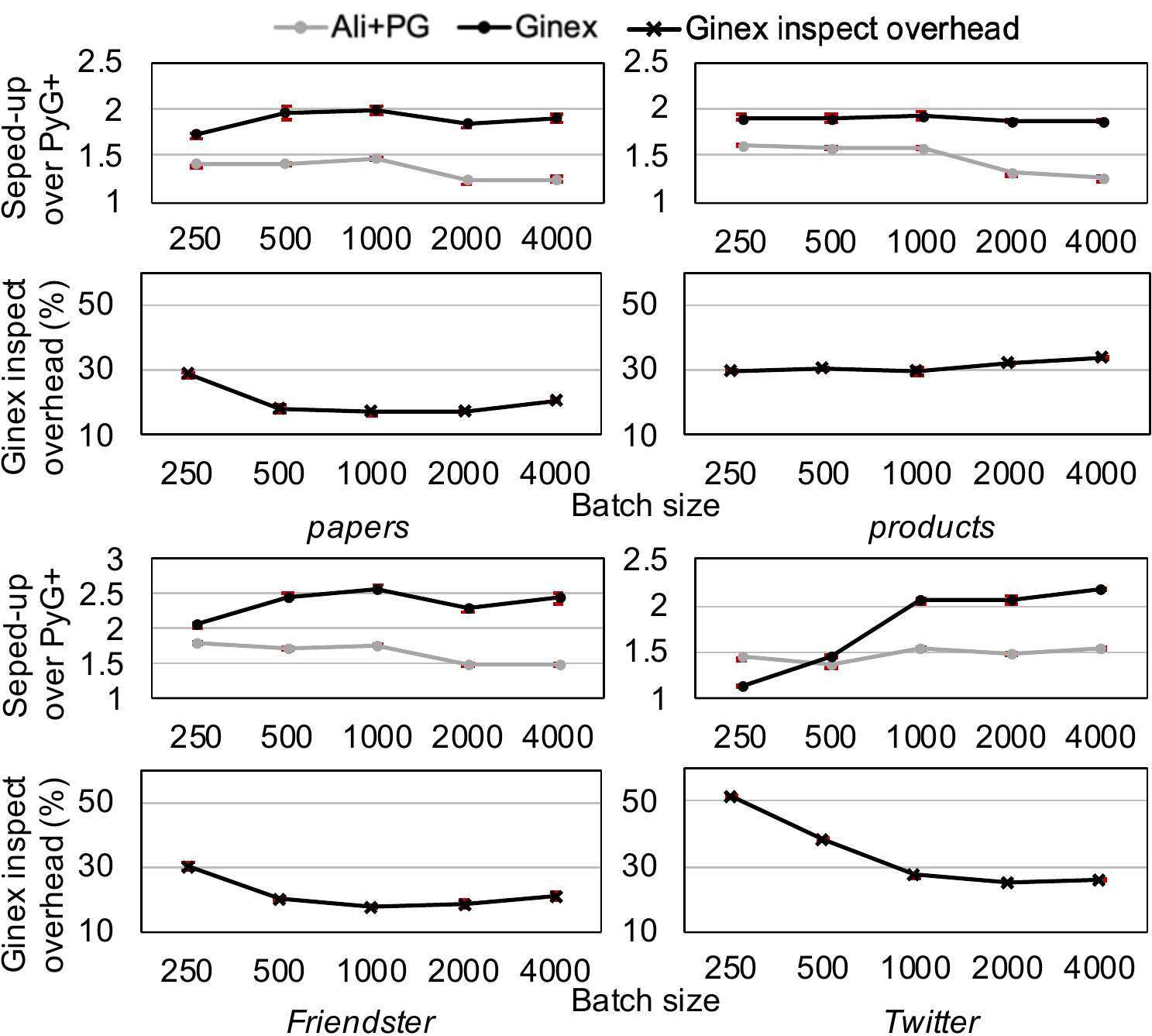}
  \caption{\rev{Speedup of Ali+PG and Ginex over PyG+ with varying batch size}}
  \label{fig:eval:batch-size}
\end{figure}
Figure~\ref{fig:eval:batch-size} shows the impact of batch
size. \name is consistently faster than Ali+PG except when batch size is very small (e.g., 250). This is because the changeset precomputation overhead of \name can be a new bottleneck at a small batch size. As the batch size decreases, the time for changeset computation tends to reduce much more slowly than the other stages which scale almost linearly. Thus, beyond a certain point, the changeset precomputation time is no longer hidden by the superbatch sample stage. However, this behavior manifests at a very low batch size outside the typical range. It is common to use larger batch sizes ($\geq$ 1000) in GNN training~\cite{batch-size} for both throughput and accuracy. 

\hypertarget{sec:eval:distributed}{}
\subsection{\rev{Cost Comparison with Distributed Training}}
\label{sec:eval:distributed}
\begin{table}[!t]
\setlength\doublerulesep{0.7pt}
\caption{\rev{Machine configurations on Google Cloud}} \label{tbl:gcp_config}
{\tablesize
\begin{tabular}{lll}
\hhline{===}
\multicolumn{3}{c}{\textbf{DistDGL}}    \\ \hhline{===}
                                           &  \multicolumn{1}{c}{Node 0}               &  \multicolumn{1}{c}{Node 1-7}                                                                                                                               \\ \hhline{---}
Host Processor                             & 24 vCPU              & 24 vCPU                                                                                                                                \\
Host Memory                                & 256 GB               & 156 GB for \emph{Friendster}  \\
                                &                & 128 GB for \emph{Twitter} \\
Host Storage                               & 2 TB pd-standard     & 100 GB pd-standard                                                                                                                     \\
GPU                                        & NVIDIA T4 $\times$ 2 & NVIDIA T4 $\times$ 2                                                                                                                   \\
Network                                    & 16 Gbps outbound BW  & 16 Gbps outbound BW                                                                                                                    \\ \hline
Hourly Cost (8 Nodes)                      & \multicolumn{2}{l}{\$18.83 for \emph{Friendster} / \$17.94 for \emph{Twitter}}                                          \\ 
\hhline{===}
\multicolumn{3}{c}{\textbf{Ginex}}                                                                                                                                                                                  \\ \hhline{===}
Host Processor & \multicolumn{2}{l}{12 vCPU}                                                                                                                                   \\
Host Memory                                & \multicolumn{2}{l}{64 GB}                                                                                                                                     \\
Host Storage                               & \multicolumn{2}{l}{375 GB NVMe SSD $\times$ 16}                                                                                                               \\
                                           & \multicolumn{2}{l}{(Seq. write: 3.12 GB/s, Seq. Read: 6.24 GB/s)}                                                                                             \\
GPU                                        & \multicolumn{2}{l}{NVIDIA T4 $\times$ 1}                                                                                                                      \\ \hline
Hourly Cost (1 Node)                       & \multicolumn{2}{l}{\$1.36} \\ \hline                                                                                                                                   
\end{tabular}%
}
\end{table}
\begingroup
\renewcommand*{\arraystretch}{1.15}
\begin{table}[!t]
\centering
\caption{\rev{Cost comparison of \name and DistDGL}} \label{tbl:cost}
{\tablesize
\begin{tabular}{P{1.5cm} P{1.5cm} P{1.9cm} P{1.9cm}}\Xhline{1.5\arrayrulewidth}
Dataset & Hourly Cost Reduction & Normalized Performance & Normalized Performance/\$ \\ \Xhline{1.5\arrayrulewidth}
\textit{Friendster} & 13.80$\times$ & 0.20$\times$ & 2.76$\times$ \\ 
\textit{Twitter} & 13.15$\times$ & 0.43$\times$ & 5.71$\times$ \\ 
\Xhline{1.5\arrayrulewidth}
\end{tabular}
}
\end{table}
\endgroup

This section presents a case study of comparing the cost efficiency of \name with that of DistDGL~\cite{dist-dgl}, a popular distributed GNN training system. DistDGL partitions the graph dataset with the min-cut partitioning algorithm~\cite{metis} and keeps it in memory over a distributed cluster. We set up both \name and DistDGL on Google Cloud and report the cost efficiency in terms of performance per dollar on GraphSAGE. We use two datasets, \textit{Friendster} and \textit{Twitter}.

Table~\ref{tbl:gcp_config} shows the configurations for DistDGL and \name. For DistDGL, we use an 8-node cluster. Each node has two NVIDIA T4 GPUs. This yields a slightly better cost-efficiency than a single-GPU setting. Since Node 0 serves as a master node and thus requires more memory than the others, we equip this node with a larger memory. We have carefully tuned the memory size for the master and worker nodes. Note that the aggregated memory size of the cluster is larger than the dataset size in Table~\ref{tbl:datasets}. This is because some parts of the graph data are duplicated over multiple nodes for performance, and managing a distributed store consumes additional memory. For \name, we use a single node with 64 GB memory, one NVIDIA T4 GPU, and 16 375 GB NVMe SSDs configured with RAID-0.

\rev{
Table~\ref{tbl:cost} reports the cost efficiency. \name achieves 2.76$\times$ and 5.71$\times$ higher cost efficiency than DistDGL in terms of performance per dollar for \textit{Friendster} and \textit{Twitter}, respectively. Although the raw training throughput of \name is lower than DistDGL, the hourly cost of \name is over 13$\times$ lower than DistDGL to make it much more cost effective. The overhead of data distribution over multiple nodes in DistDGL incurs a significant cost when scaling GNN training.  }

\section{Related Work}
\label{sec:related}

\para{Scalable Graph Neural Networks}
To the best of our knowledge, \name is the first to leverage commodity SSDs to scale GNN training. GLIST~\cite{glist2021li} also utilizes SSDs to scale GNN but focuses on inference and requires specialized hardware. Instead, most proposals for large-scale GNN training have taken \emph{scale-out} approaches. ROC~\cite{roc} and NeuGraph~\cite{neugraph} propose multi-GPU training system for GNN. However, they adopt full-batch training which makes them eventually face the GPU memory capacity wall when training on very large graphs. Meanwhie, distributed systems utilizing multiple CPU nodes for graph storage present more scalable options~\cite{p32021ghandi,aligraph2019zhu,agl,cm-gcn,dist-dgl,dist-dglv2}. Although details vary, they partition the graph dataset and keep it in memory on a cluster. However, a surge of system cost limits cost-effectiveness of this approach.


\para{Caching Mechanism for Graph Processing}
There are proposals for caching mechanisms specially designed for graph processing. GRASP~\cite{GRASP} classifies nodes into three categories according to their degrees and manages nodes in different categories with different caching policies. Graphfire~\cite{graphfire} manages cached data in a fine-grained manner by learning access patterns online with a locality predictor. However, their caching performance is sub-optimal, and it takes substantial effort to implement them in software. P-OPT~\cite{p-opt} uses the transpose of a graph's adjacency matrix to closely mimic the optimal caching policy. However, it assumes deterministic graph traversal patterns, and hence is not applicable to GNN training, where the seed nodes for each mini-batch are randomly selected and the sampling process might also require some randomization.

\para{Scaling-up of Other Graph Workloads}
There have been many proposals to optimize disk-based large-scale graph processing on a single node~\cite{marius2021mohoney, mosaic2017maass, gts2016kim, flashgraph2015zheng, gridgraph2015zhu, graphchi2012kyrola, xstream}. GraphChi~\cite{graphchi2012kyrola} makes various graph workloads available on a PC by a parallel sliding windows method. X-stream~\cite{xstream} reduces the disk accesses by an edge-centric approach. FlashGraph~\cite{flashgraph2015zheng} and MOSAIC~\cite{mosaic2017maass} adopt a custom data structure for graph. Marius~\cite{marius2021mohoney} optimizes graph embedding learning by partition caching and buffer-aware data orderings. While these works share the same spirit with \name to replace or augment a cluster-based approach with storage devices, they target non-GNN workloads, which have different characteristics.


\para{Scaling DNN Training with SSD}
Several proposals use SSD to scale training of a large-scale DNN other than GNN~\cite{dragon,flashneuron,behemoth}.  Dragon~\cite{dragon} and FlashNeuron~\cite{flashneuron} leverage direct storage access as a backing store that augments GPU memory. Behemoth~\cite{behemoth} introduces DNN training accelerator which replaces HBM DRAM with flash memory. While their goal is to overcome GPU memory capacity wall which stems from the excessive size of model or intermediate data, \name focuses on the CPU memory capacity wall in GNN training, which is required to process much larger datasets.

\section{Conclusion}
\label{sec:conc}
Training GNNs are often challenging as real-world graph datasets are usually very large exceeding the main memory capacity. Distributed training is an option, but may not be cost-effective. Thus, we propose \name, an SSD-based GNN training system that supports billion-scale graph datasets on a single machine. By reconstructing the training pipeline, \name realizes the optimal caching system for feature vectors thus alleviating the I/O bottleneck. With the reduced I/O overhead, \name can scale GNN training to datasets an order-of-magnitude larger than a single machine's CPU memory capacity, while achieving substantial speedups over the state-of-the-art.


\begin{acks}
This work was supported by SNU-SK Hynix Solution Research Center (S3RC) and the National Research Foundation of Korea (NRF) grant funded by Korea government (MSIT) (NRF-2020R1A2C3010663). Jae W. Lee is the corresponding author.

\end{acks}


\balance
\bibliographystyle{ACM-Reference-Format}
\bibliography{references}

\end{document}